\documentclass[10pt,twocolumn,letterpaper]{article}

\usepackage{iccv}
\usepackage{times}
\usepackage{epsfig}
\usepackage{graphicx}
\usepackage{amsmath}
\usepackage{amssymb}
\usepackage{helvet}
\usepackage{courier}
\usepackage{amsmath,amssymb,amsfonts}
\usepackage{algorithm,algorithmic}
\usepackage{color}
\usepackage{graphicx}
\usepackage{subfigure}
\usepackage{multirow}
\usepackage{enumerate}
\usepackage{multirow}
\usepackage{marvosym}

\newenvironment{proof}[1][Proof]{\begin{trivlist}
\item[\hskip \labelsep {\bfseries #1}]}{\end{trivlist}}
\newcommand{\invisible}[1]{{}}
\newtheorem{theorem}{Theorem}
\newtheorem{lemma}{Lemma}
\newtheorem{prop}{Proposition}

\newtheorem{assumption}{Assumption}
\newtheorem{remark}{Remark}
\newtheorem{criterion}{Criterion}

\newcommand{\X}{\mathbf{X}}
\newcommand{\x}{\mathbf{x}}
\newcommand{\y}{\mathbf{y}}

\newcommand{\sv}{\mathbf{v}}
\newcommand{\z}{\mathbf{z}}
\newcommand{\I}{\mathbf{I}}

\newcommand{\D}{\mathbf{D}}
\newcommand{\w}{\mathbf{w}}

\newcommand{\W}{\mathbf{W}}

\newcommand{\e}{\mathbf{e}}
\newcommand{\bd}{\mathbf{d}}

\usepackage[pagebackref=true,breaklinks=true,letterpaper=true,colorlinks,bookmarks=false]{hyperref}

\iccvfinalcopy % *** Uncomment this line for the final submission

\def\iccvPaperID{1536} % *** Enter the ICCV Paper ID here

\ificcvfinal\fi
\begin{document}

%%%%%%%%% TITLE
\title{An Optimization Framework with Flexible Inexact Inner Iterations \\ for Nonconvex and Nonsmooth Programming}

\author{Yiyang Wang\\
School of Mathematical Sciences\\
Dalian University of Technology\\
{\tt\small yywerica@gmail.com}
\and
Risheng Liu\\
School of Software Technology\\
Dalian University of Technology\\
{\tt\small rsliu@dlut.edu.cn}
\and
Xiaoliang Song\\
School of Mathematical Sciences\\
Dalian University of Technology\\
{\tt\small ericsong507@gmail.com}
\and
Zhixun Su\\
School of Mathematical Sciences\\
Dalian University of Technology\\
{\tt\small zxsu@dlut.edu.cn}
}

\maketitle
%\thispagestyle{empty}

%%%%%%%%% ABSTRACT
\begin{abstract}
In recent years, numerous vision and learning tasks have been (re)formulated as nonconvex and nonsmooth programmings (NNPs).
Although some algorithms have been proposed for particular problems, designing fast and flexible optimization schemes with theoretical guarantee is a challenging task for general NNPs.
It has been investigated that performing inexact inner iterations often benefit to special applications case by case, but their convergence behaviors are still unclear.
Motivated by these practical experiences, this paper designs a novel algorithmic framework, named inexact proximal alternating direction method (IPAD) for solving general NNPs.
We demonstrate that any numerical algorithms can be incorporated into IPAD for solving subproblems and the convergence of the resulting hybrid schemes can be consistently guaranteed by a series of simple error conditions.
Beyond the guarantee in theory, numerical experiments on both synthesized and real-world data further demonstrate the superiority and flexibility of our IPAD framework for practical use.
\end{abstract}

%%%%%%%%% BODY TEXT
\section{Introduction}\label{Sec:Intro}
%%%%%%%%%%%%%%%%%%%%%%%%%%%%%%%%%%%%%%%%%%%介绍我们这篇文章关注的模型的一般形式

Nonconvex and nonsmooth programmings (NNPs) have received wide attentions in recent years~\cite{Shi2011A,Kim2011Fast,xu2013block,yuan2013l0tv,bao2016dictionary}.
Many problems in vision and learning societies, such as sparse coding and dictionary learning~\cite{bao2016dictionary}, matrix factorization~\cite{Kim2011Fast}, image restoration~\cite{yuan2013l0tv}, and image classification~\cite{Zou2012Sparse} can be (re)formulated as specific NNPs.
In this work, we consider a general NNP in following formulation:
\begin{equation}
\min_{\X:=(\x_1,\ldots, \x_K)}\Psi(\X):= \sum\limits_{i=1}^K f_i(\x_i) + H(\X),\label{eq:primal}
\end{equation}
where $\x_i$s are vectors or matrices throughout the paper.

This general NNP covers a variety of techniques in image processing and machine learning.
For example, principal component analysis with sparse constraints, like lasso constraint \cite{Shen2008Sparse} and elastic-net regularization \cite{Zou2012Sparse} can be written in problem (\ref{eq:primal}).
In non-negative matrix factorization problem \cite{Kim2011Fast}, it is common to adopt $H$ as a $\ell_2$ distance on describing the ability of restoration and restrict $f_i$s to be non-negative on each component.
Sparse dictionary learning (SDL) can also be formulated in problem (\ref{eq:primal}).
%and its two factorized components are called ``dictionary'' and ``codes'', respectively.
Many literatures \cite{Gregor2010Learning,Shi2011A,bao2016dictionary} have posted the superiority on restricting dictionaries with normalized bases while at the same time constraining sparsity for codes with various nonconvex and nonsmooth sparse penalties.
%Since SDL is a powerful tool to represent diverse kinds of data in image processing and machine learning \cite{Gregor2010Learning}, its algorithms and applications have gained wider attentions and further discussions.

%%%%%%%%%%%%%%%%%%%%%%%%%%%%%%%%%%%%%%%%%%针对这样的一般问题，已经提出了一些有收敛性的算法。
In the past few years, there have been literatures \cite{attouch2010proximal,bolte2014proximal,xu2013block,xu2014globally,Bolte2015Majorization} in optimization and numerical analysis on designing converged algorithms in view of the general NNP (\ref{eq:primal}).
These algorithms have been applied to various problems, such as non-negative matrix factorization \cite{bolte2014proximal}, SDL with $\ell_0$ penalty \cite{bao2016dictionary} and non-negative Tucker decomposition \cite{xu2014globally}; at the same time, abundant experimental analyses have demonstrated the efficiency and convergence properties of these algorithms.
However, in pursuit of convergence, most of the existing algorithms are designed with fixed iteration schemes; those inflexible schemes are tightly constrained and fail to take the model structures of specific problems into consideration.

%%%%%%%%%%%%%%%%%%%%%%%%%%%%%%%%%%%%%%%%%%
While on the other hand, specific solvers designed for practical use are far more flexible than the converged algorithms proposed for the general problem (\ref{eq:primal}).
Those specific solvers always take advantages of the problem structures and then employ effective numerical methods to solve specific sub-problems.
Moreover, we have noticed from previous work \cite{bresson2009short,wang2016nonlocal,guo2014robust,Chen2016Dehazing,Pan2017} that solving sub-problems with inner iterations is a frequently used strategy in numerous efficient solvers.
Though few of the inexact solvers are designed with rigid theoretical support and analyses, their efficiencies and convergence properties have been verified from experiments under certain conditions.

\subsection{Contributions}
%%%%%%%%%%%%%%%%%%%%%%%%%%%%%%%%%%%%%%%%%%我们文章的贡献
Motivated from various inexact solvers, we in this paper propose an unified and flexible algorithm framework named \emph{inexact proximal alternating direction} method (IPAD).
Our IPAD is designed for solving the general NNP problem (\ref{eq:primal}), at the same time, keeping the flexibility when dealing with specific problems.
Different from existing solvers in practice, IPAD theoretically give rigid conditions on parameters and stopping criteria to ensure the convergence, thus is more rigid and robust for practical use.
The theoretical support provided in this paper can be regarded as a guidance for designing inexact algorithms for solving specific problems in a concise framework.
As far as we know, we are the first to incorporate various numerical methods into a general algorithm framework and at the same time give rigorous convergence analyses for NNPs.
In summary, our contributions are three folds:
\begin{enumerate}
\item Different from most existing numerical algorithms for NNPs, which always fix their updating schemes
during iterations, we provide a novel perspective to incorporate different optimization strategies into a unified and flexible proximal optimization framework
for \eqref{eq:primal}.

\item Even with inexact subproblems and flexible inner iteration schemes,
we prove that the convergence property of the resulting hybrid optimization framework can still be guaranteed. Indeed, our theoretical results are the best we can ask for, unless further assumptions are made on the general NNPs in \eqref{eq:primal}.

\item
As an application example, we show implementations of applying IPAD with different inner algorithms to solve
the widely concerned $\ell_0$ regularized SDL model.
Numerical evaluations and abundant experimental comparisons demonstrate promising experimental results of the proposed algorithms, which verify the flexibility and efficiency of our optimization framework.
	
\end{enumerate}

%\begin{enumerate}
%\item
%We propose a general algorithm framework named IPAD for solving the non-convex and non-smooth problem (\ref{eq:primal}).
%Our proposed IPAD is not a single algorithm, but a flexible algorithm framework.
%That is, it allows fusing various effective numerical methods into, which helps improve the efficiency of the whole algorithm.
%
%\item
%By restricting conditions on parameters and stopping criteria, our IPAD is proved to achieve the best convergence result in general non-convex and non-smooth problems.
%In addition, we also provide a truly implementable process for carrying our proposed IPAD.
%
%\item
%We apply IPAD to a widely concerned SDL problem with $\ell_0$ penalty, on both synthetic and real-world data; the abundant experimental comparisons verify the flexibility and efficiency of our IPAD.
%\end{enumerate}

\section{Related Work}\label{Sec:Rela_Work}

\subsection{Existing Optimization Strategies for NNPs}\label{Sec:2-1}

In past several years, accompanied with the rising popularity of investigating sparsity and low-rankness (naturally with nonconvex and nonsmooth properties) for vision and learning tasks~\cite{Kim2011Fast,yuan2013l0tv,Xu2013Unnatural,Afonso2015Blind,Lu2015Nonconvex,Zhang2015Robust}, developing numerical solvers for different types of NNP models have attracted considerable research interests from computer vision, machine learning and optimization societies.

As a typical nonconvex and nonsmooth measurement, $\ell_0$ norm has been widely applied for image processing problems \cite{yuan2013l0tv,Xu2013Unnatural,Afonso2015Blind};
and various methods have been designed for solving its optimization.
In \cite{yuan2013l0tv}, the authors propose a proximal method for finding desirable solution to $\ell_0$TV problem, but they only prove a weak convergence result under mild conditions.
The rank function \cite{Lu2015Nonconvex,Zhang2015Robust} has also been well studied: the work in \cite{Lu2015Nonconvex} optimizes a low-rank minimization by an iteratively reweighted algorithm.
However, they can only prove that any limit point is a stationary point, which cannot guarantee the correct convergence of the iterations.
Another typical problem, non-negative matrix factorization is a powerful technique for various applications; algorithms like ANLS \cite{Kim2011Fast} have been specially designed for solving this specific problem.

On the other hand, some algorithms with rigid theoretical analyses have been proposed for solving the general NNP problem \eqref{eq:primal} \cite{attouch2010proximal,bolte2014proximal,Xu2016Relaxed}.
The authors in \cite{attouch2010proximal} propose an algorithm named GIST for solving a special class of NNP \eqref{eq:primal} and apply it to logistic regression problem.
A proximal alternating linearized method is proposed in \cite{bolte2014proximal}, however, it is time-consuming for solving special problems.
Xu et al. in \cite{Xu2016Relaxed} propose a MM method for the general NNP, but their algorithm is too complicated to be implemented in practice.
The authors in \cite{attouch2010proximal} provide their convergence result by assuming subproblems are exactly solved, however, this condition is unattainable in most situations.

In summary, most optimization schemes for specific NNPs are efficient in practice, however, their convergence guarantees are relatively weak.
While, algorithms designed for the general NNP are well-converged, but they always have strict conditions and are inefficient in practice.

\subsection{Inexact Inner Iterations in Application Fields}\label{Sec:2-2}

We have observed that in some application fields, relatively good performances are often obtained by ``improper'' numerical algorithms.
%That is, optimization with inexact inner iterations may lead to good results for solving subproblems in specific tasks.
In previous literatures \cite{Zuo2016Learning,Yang2016admmnet,Pan2017}, the ``improper'' iteration skills have been frequently used in solving sparse coding, blind kernel estimation, and medical imaging problems in practice.

One of those inexact schemes is employing numerical methods as inner iteration methods for solving special subproblems.
For instance, the authors in \cite{Zuo2016Learning} employ half-quadratic splitting and Lagrangian dual method as the inner iteration schemes for estimating clear images in blind deconvolution problem.
While in a text image deblurring paper \cite{Pan2017}, the authors update variables though alternating direction method, meanwhile, applying half-quadratic splitting for solving subproblems.
Although these inexact solvers are effective in practice, they are totally designed in freestyle and are uncontrolled in the performances.

Another kind of inexact schemes is adopting neural networks for approximating exact solutions during iterations.
For example, time-unfolded recurrent neural networks can be used to produce the best possible approximations for sparse coding problem \cite{Gregor2010Learning,Sprechmann2012Learning}.
On the other hand, fusing various neural networks in ADMM framework \cite{Chen2015Trainable,Yang2016admmnet} has great performances for approximately optimizing problems for image restoration and compressed sensing.
The efficiencies of these inexact solvers and their convergence performances have been verified from experiments, however, few of them are designed with rigid theoretical support.

%%%%%%%%%%%%%%%%%%%%%%%%%%%%%%%%%%%%%%%%%%%%%%%%%%%%%%%%%%%%%%%%%%
%%%%%%%%%%%%%%%%%%%%%%%%%%%%%%%%%%%%%%%%%%%%%%%%%%%%%%%%%%%%%%%%%%
\section{The Proposed Optimization Framework}

To simplify subsequent derivations, we propose an IPAD and analyze its properties for problem (\ref{eq:primal}) with $2$ variables\footnote{For simplicity, we replace the notations in problem (\ref{eq:primal}) with the new ones. We hope this will not cause confusions.}:
\begin{equation}\label{eq:2variable}
\min\limits_{\z:=(\x,\y)}\Psi(\z):= f(\x) + g(\y) + H(\x, \y).
\end{equation}
Though the whole properties are analyzed for problem (\ref{eq:2variable}), it is straightforward to extend them to the general one.
Then we give assumptions on the objective function:

\noindent (1) $f$ and $g$ are proper, lower semi-continuous functions;

\noindent (2) $H$ is a $C^1$ function and its gradient is Lipschitz continuous on a bounded set;

\noindent (3) $\Psi$ is a coercive, Kurdyka-{\L}ojasiewicz (K{\L}) function\footnote{To be self-contained, the definition of K{\L} function is given in the supplemental material due to space limit.}.

\begin{remark}
It should be mentioned that the most frequently used $\ell_2$-norm $H(\x,\y)$ is a K{\L} function which also satisfies the second assumption.
On the other hand, regularizers like $\ell_0$ penalty, $\ell_p$ norm, SCAD \cite{fan2001variable}, MCP \cite{zhang2010nearly} and indicator functions are all K{\L} functions and at the same time satisfy the first assumption.
Since the finite sums of K{\L} functions are also K{\L} functions \cite{bolte2014proximal}, thus not a few models in image processing and machine learning satisfy these assumptions.
\end{remark}

\subsection{Inexact Proximal Reformulation}

For solving problem (\ref{eq:2variable}), it is natural and common to apply a proximal alternating direction method (PAD) \cite{xu2012alternating,Chen2016Dehazing} for alternatively updating variables in a cyclic order:
\begin{equation}\label{iPAM:ex_sequence}
\begin{aligned}
\x^{t}_{\ast} &= \arg\min\limits_{\x} f(\x) + H(\x, \y^{t-1}) + \frac{\eta_1^{t-1}}{2}\|\x-\x^{t-1}\|^2,\\
\y^{t}_{\ast} &= \arg\min\limits_{\y} g(\y) + H(\x^{t}, \y) + \frac{\eta_2^{t-1}}{2}\|\y-\y^{t-1}\|^2,
\end{aligned}
\end{equation}
where $\eta_1^{t-1}$ and $\eta_2^{t-1}$ are proximal parameters added on the subproblems respectively; the notations $\x^{t}_{\ast}$ and $\y^{t}_{\ast}$ are used to represent exact solutions of the subproblems in Eq. (\ref{iPAM:ex_sequence}).
It should be pointed out that adding proximal terms on subproblems is a common skill to improve the stability of algorithms on solving nonconvex problems \cite{attouch2010proximal}.

As claimed before, in most cases, it is either impossible or extremely hard for calculating \emph{exact} solutions of subproblems.
Thus not a few work employ inner iteration schemes to compute approximations to the exact solutions, which means, the \emph{inexact} solutions $\x^{t}$ and $\y^{t}$ calculated by inner methods are approximations to $\x^{t}_{\ast}$ and $\y^{t}_{\ast}$:
\begin{equation}\label{iPAM:iex_sequence}
\x^{t} \approx \x^{t}_{\ast}, \quad \quad \y^{t} \approx \y^{t}_{\ast}.
\end{equation}

We can see from the above that our IPAD is in a general framework: it alternatively updates variables by approximately solving the subproblems of PAD.
While on the other hand, it does not restrict specific formulas for solving the subproblems, hence IPAD has quite flexible inner iteration schemes that can fuse efficient numerical methods into.

%\begin{algorithm}[!t]
%\caption{Inexact Proximal Alternating Direction}
%\begin{algorithmic}[1]
%\STATE Setting parameters: $C_x$, $C_y$, $\{\eta_1^t\}_{t\in \mathbb{N}}$, $\{\eta_2^t\}_{t \in \mathbb{N}}$.
%\STATE Initializing variables: $\x^0$, $\y^0$.
%\WHILE {Not Converged}
%\STATE $\x^{t} = \psi(\x; \x^{t-1}, \y^{t-1}, f, H, \eta_1^t, C_x)$.
%\STATE $\y^{t} = \psi(\y; \y^{t-1}, \x^{t}, g, H, \eta_2^t, C_y)$.
%\ENDWHILE
%\STATE (Function $\psi(\mathbf{u}; \mathbf{u}^{t-1}, \mathbf{v}, h, H, \eta, C_u)$ with respect to $\mathbf{u}$ is denoted as inner iteration schemes in Alg. \ref{Alg:iPAM2}.)
%\end{algorithmic}\label{Alg:iPAM1}
%\end{algorithm}

\begin{algorithm}[!t]
\caption{Inexact Proximal Alternating Direction}
\begin{algorithmic}[1]
\STATE Setting parameters: $C_x$, $C_y$, $\{\eta_1^t\}_{t\in \mathbb{N}}$, $\{\eta_2^t\}_{t \in \mathbb{N}}$.
\STATE Initializing variables: $\x^0$, $\y^0$.
\WHILE {not converged}
\STATE $\x^{t} = \psi(\x; \x^{t-1}, \y^{t-1}, f, H, \eta_1^t, C_x)$. \\(i.e., Perform Alg. \ref{Alg:iPAM2} for $\x$ subproblem)
\STATE $\y^{t} = \psi(\y; \y^{t-1}, \x^{t}, g, H, \eta_2^t, C_y)$. \\(i.e., Perform Alg. \ref{Alg:iPAM2} for $\y$ subproblem)
\ENDWHILE
\end{algorithmic}\label{Alg:iPAM1}
\end{algorithm}

\begin{algorithm}[!t]
\caption{$\mathbf{u}^t = \psi(\mathbf{u}; \mathbf{u}^{t-1}, \mathbf{v}, h, H, \eta, C_u)$.}
\begin{algorithmic}[1]
\STATE With parameters $C_x$, $C_y$, $\{\eta_1^t\}_{t\in \mathbb{N}}$, $\{\eta_2^t\}_{t \in \mathbb{N}}$ in Alg. \ref{Alg:iPAM1}.
\STATE Denote $\mathcal{A}$ as inner iteration schemes for optimizing $h(\mathbf{u}) + H(\mathbf{u}, \mathbf{v}) + \frac{\eta}{2}\|\mathbf{u}-\mathbf{u}^{t-1}\|^2$.
\STATE Let $\mathbf{u}^{t,0}=\mathbf{u}^{t-1}$.
\WHILE {$\|\e_u^{t,i}\| > C_u\|\mathbf{u}^{t,i}-\mathbf{u}^{t-1}\|$}
\STATE $\mathbf{u}^{t,i} = \mathcal{A}(\mathbf{u}^{t, i-1})$.
\STATE In theory, use Eq. \eqref{iPAM:iex_KKT} to analyze convergence.
\STATE In practice, use Eq. \eqref{Alg_Imp:implem_err} for judging criterion.
\ENDWHILE
\STATE $\mathbf{u}^{t}=\widetilde{\mathbf{u}}^{t,i}$.
\end{algorithmic}\label{Alg:iPAM2}
\end{algorithm}

%\begin{algorithm}[!t]
%\caption{$\mathbf{u}^t = \psi(\mathbf{u}; \mathbf{u}^{t-1}, \mathbf{v}, h, H, \eta, C_u)$.}
%\begin{algorithmic}[1]
%\STATE With parameters $C_x$, $C_y$, $\{\eta_1^t\}_{t\in \mathbb{N}}$, $\{\eta_2^t\}_{t \in \mathbb{N}}$ in Alg. \ref{Alg:iPAM1}.
%\WHILE {$\|\e_u^{t}\| > C_u\|\mathbf{u}^{t}-\mathbf{u}^{t-1}\|$}
%\STATE Compute $\mathbf{u}^{t}$ by any numerical method.
%\STATE $\sv_u^{t}=\mathbf{u}^{t} - \nabla_{\mathbf{u}}H(\mathbf{u}^{t}, \mathbf{v}) - \eta(\mathbf{u}^{t}-\mathbf{u}^{t-1}).$
%\STATE $\widetilde{\mathbf{u}}^{t} = \mathrm{prox}^{1}_{h}(\sv_u^{t})$.
%\STATE $\e_u^{t}= (1-\eta)(\widetilde{\mathbf{u}}^{t}-\mathbf{u}^{t})+\nabla_{\mathbf{u}}H(\mathbf{u}^{t}, \mathbf{v})-\nabla_{\mathbf{u}}H(\widetilde{\mathbf{u}}^{t}, \mathbf{v})$.
%\ENDWHILE \ \ Re-assign $\mathbf{u}^{t}=\widetilde{\mathbf{u}}^{t}$.
%\end{algorithmic}\label{Alg:iPAM2}
%\end{algorithm}

\subsection{Flexible Inner Iteration Schemes}

Our IPAD framework is highly flexible since it allows fusing any efficient algorithms, to compute inexact solutions for specific subproblems.
This flexibility is especially welcomed in practice: since the constraints added on variables are always quite different from one another, efficient algorithms for solving specific subproblems should be carefully chosen.
For example, when facing the SDL problem with $\ell_1$ regularization \cite{elad2006image,Mairal2010Online}, both subproblems are convex.
Then various numerical methods designed under convex case like homotopy method \cite{Efron2004Least}, FISTA \cite{Beck2009A} and ADMM \cite{boyd2011distributed} can be applied to solving these subproblems.

On the other hand, although many previously used inexact solvers can be presented under our IPAD framework \cite{Huang2015A,wang2016nonlocal,Chen2016Dehazing}, almost all the inner iteration schemes are experimentally designed.
For example, a 2-step inner loop is supposed to be ``good-enough'' in \cite{wang2016nonlocal}; the authors of \cite{Huang2015A} stop the inner iterations at fixed steps to achieve acceptable solutions.
Different from the previous work, we in the following Criterion \ref{crit} provide theoretical conditions for stopping the inner iterations.

Since calculating the inexact solutions brings errors $\e_x^{t}$ and $\e_y^{t}$ in the first-order optimality conditions:
\begin{equation}\label{iPAM:iex_KKT}
\begin{aligned}
\e_x^{t} &= \mathbf{g}_x^{t} + \nabla_{\x} H(\x^{t}, \y^{t-1}) + \eta_1^{t-1}(\x^{t}-\x^{t-1}),\\
\e_y^{t} &= \mathbf{g}_y^{t} + \nabla_{\y} H(\x^{t}, \y^{t}) + \eta_2^{t-1}(\y^{t}-\y^{t-1}),
\end{aligned}
\end{equation}
with $\mathbf{g}_x^{t} \in \partial f(\x^{t})$ and $\mathbf{g}_y^{t} \in \partial g(\y^{t})$, we in the following Criterion \ref{crit} show that these errors should be bounded to a certain extent at every iteration.

\begin{criterion}\label{crit}
The errors $\e_x^t$ and $\e_y^t$ in Eq. (\ref{iPAM:iex_KKT}) must satisfy
\begin{equation}\label{Alg_Imp:err_crit}
%\begin{aligned}
\|\e_x^{t}\| \leq C_x\|\x^{t}-\x^{t-1}\|,\ \  \|\e_y^{t}\| \leq C_y\|\y^{t}-\y^{t-1}\|,
%\end{aligned}
\end{equation}
where parameters $C_x$ and $C_y$ are two positive integers defined before the iteration starts.
\end{criterion}
%
%We in this paper focus on providing conditions on parameters and inner stopping criteria to ensure the convergence of this inexact algorithm framework.
%The theoretical support in this paper will guide the design of inexact solvers for specific problems in applications.
%Being as the first inexact method for NNP problems (\ref{eq:primal}), IPAD is \emph{practicable, converged} and \emph{efficient}.
%We in the next sub-section present the \emph{practicability} and then analyze the \emph{convergence} in Sec. \ref{Sec:xxx}.
%The \emph{efficiency} of iPAM is confirmed by experimental results in Sec. \ref{Sec:Experiments}.

With the stopping criteria (\ref{Alg_Imp:err_crit}) for inner iteration schemes, we summarize the whole algorithm in Alg. \ref{Alg:iPAM1}.
%which is not a single algorithm, but an algorithm framework that can fuse any efficient numerical methods into.
However, this inexact framework is still on conceptual progress: the stopping criteria can not be directly calculated from \eqref{Alg_Imp:err_crit} and an implementation should be put forward for carrying IPAD for practical use.
Before providing a practical implementation in Sec. \ref{Sec:imp}, we first analyze the convergence properties of IPAD with the help of the well-designed Criterion \ref{crit}.

%%%%%%%%%%%%%%%%%%%%%%%%%%%%%%%%%%%%%%%%%%%%%%%%%%%%%%%%%%%%%%%%%%%%%%%%%%%%%%%%%%%%%%%%%
\subsection{Convergence Analyses}\label{Sec:xxx}

In this section, we provide the theoretical support for IPAD method\footnote{Due to space limit, all the related proofs in this section will be detailedly given in the supplemental material.}.
The strategic point on analyzing the convergence properties of IPAD is regarding $\x^{t+1}$ and $\y^{t+1}$ as the \emph{exact} solutions on solving the following subproblems:
\begin{equation}\label{Converg_Ana:inexact_exact}\small
\begin{aligned}
&\min\limits_{\x} f(\x) + H(\x, \y^t) + \frac{\eta_1^t}{2}\|\x-\x^t\|^2 - (\e_x^{t+1})^{\top}\x,\\
&\min\limits_{\y} g(\y) + H(\x^{t+1}, \y) + \frac{\eta_2^t}{2}\|\y-\y^t\|^2 - (\e_y^{t+1})^{\top}\y.
\end{aligned}
\end{equation}
This equivalent conversion is rigid since the first-order optimality conditions of (\ref{Converg_Ana:inexact_exact}) are exactly the same with Eq. (\ref{iPAM:iex_KKT}).
However, it should be emphasized that \emph{$\x^{t+1}$ and $\y^{t+1}$ are not computed by directly minimizing (\ref{Converg_Ana:inexact_exact});
this equivalent conversion is nothing but assisting in theoretical analyses.}

Before proposing the main theorem, we give the requirements on $\{\eta_1^t\}_{t\in \mathbb{N}}$ and $\{\eta_2^t\}_{t\in \mathbb{N}}$.
\begin{assumption}\label{para:ASS}
Parameters $\{\eta_1^t\}_{t\in \mathbb{N}}$ and $\{\eta_2^t\}_{t\in \mathbb{N}}$ satisfy
\begin{equation}\label{Converg_Ana:assump_etas}
\eta_1^t > 2C_x, \quad \eta_2^t > 2C_y, \quad {\rm for \ all\  t\in \mathbb{N}},
\end{equation}
to ensure the whole IPAD algorithm converges.
\end{assumption}

%Together with Criterion \ref{crit} and Eq. \eqref{iPAM:iex_sequence}, we can see from Assumption \ref{para:ASS} that if $C_x$ and $C_y$ are too large, then it relatively reduce the inner iterations but at the same time converges with a tiny step size.
%On the other side, if $C_x$ and $C_y$ are too small, then it converges with relatively large step size but will increase the inner iterations.
%Thus, these parameters are balances that should be carefully chosen.

Then with the help of Assumption \ref{para:ASS}, we can obtain the main theorem in Theorem \ref{Converg_Ana:main_theorem}: our proposed IPAD has the best convergence property, that is, the global convergence property for general NNP: \emph{our IPAD generates a Cauchy sequence that converges to a critical point of the problem.}
\begin{theorem}\label{Converg_Ana:main_theorem}
Under the Assumption \ref{para:ASS} and suppose the sequence $\{(\x^t, \y^t)\}_{t\in \mathbb{N}}$ generated by IPAD is bounded.
Then $\{(\x^t, \y^t)\}_{t\in \mathbb{N}}$ is a Cauchy sequence that converges to a critical point $(\x^{\ast}, \y^{\ast})$ of $\Psi$.
\end{theorem}

For proving this main convergence theorem, the two assertions in the following key lemma are the cornerstones.
From the two assertions, obviously, the objective function $\Psi$ is sufficiently descent (Eq. (\ref{Converg_Ana:Eq_suffDesecnt}) in Lemma \ref{Converg_Ana:key_lemma}) during iterations, which together with the second assertion ensures that there is a subsequence of $\{\z^t\}_{t\in\mathbb{N}}$ that converges to a critical point of the problem.
By further combining the K{\L} property, we have the global convergence property for our proposed IPAD method as claimed in Theorem \ref{Converg_Ana:main_theorem}.

\begin{lemma}\label{Converg_Ana:key_lemma}
Suppose that the sequence $\{(\x^t, \y^t)\}_{t\in \mathbb{N}}$ generated by IPAD is bounded.
Then the following two assertions hold under the Assumption \ref{para:ASS}:
\begin{equation}\label{Converg_Ana:Eq_suffDesecnt}\small
\Psi(\z^t) - \Psi(\z^{t+1})\geq a(\|\x^{t+1}-\x^t\|^2 + \|\y^{t+1}-\y^t\|^2),
\end{equation}
\begin{equation}\label{Converg_Ana:Eq_subgradient}\small
dist(0, \partial \Psi(\z^t)) \leq b(\|\x^t - \x^{t-1}\| + \|\y^t-\y^{t-1}\|),
\end{equation}
where constants $a=\min_{t\in \mathbb{N}}\{\frac{\eta_1^t}{4}-\frac{(C_x)^2}{\eta_1^t}, \frac{\eta_2^t}{4}-\frac{(C_y)^2}{\eta_2^t}\}$ and $b = \max_{t\in \mathbb{N}}\{\eta_1^{t-1}+C_x, M+\eta_2^{t-1}+C_y\}$ with Lipschitz constant $M$ of $\nabla H$ on bounded sets.
\end{lemma}

For the convergence rate, our IPAD method shares the same result with PALM \cite{bolte2014proximal} when the desingularising function $\phi (s)=\frac{C}{\theta}s^{\theta}:[0,\mu)\to \mathbb{R}_+$ of $\Psi$ is satisfied with positive constant $C$ and $\theta \in [0, 1).$
Specifically, IPAD converges in a finite number of steps when $\theta$ equals to 1.
For $\theta \in [0, \frac{1}{2})$ and $\theta \in (\frac{1}{2}, 1)$, IPAD converges with a sublinear rate and a linear rate respectively.
Though the convergence rate of IPAD is the same with PALM in theory (the convergence rate is not affected by the algorithm but the objective function $\Psi$ of NNP \cite{Frankel2015Splitting}), the experimental results in Section \ref{Sec:Experiments} verify the efficiency of our algorithm.

\begin{remark}
It is natural to extend IPAD to solving the general multi-block NNP \eqref{eq:primal}.
Furthermore, all the convergence analyses conducted on problem \eqref{eq:2variable} can be straightforwardly extended to the multi-block case.
Since not a few problems can be (re)formulated as the general formulation \eqref{eq:primal}, thus our IPAD method can be applied for solving a wider range of problems in applications.
\end{remark}

We can see from the above contents that, arbitrarily incorporating inner numerical methods into IPAD still ensure the global convergence property of the whole algorithm as long as the stopping criteria \eqref{Alg_Imp:err_crit} and the parameter conditions (\ref{Converg_Ana:assump_etas}) are satisfied.
From this perspective, IPAD is not a single algorithm, but a general and flexible algorithm framework with rigid convergence properties.
In addition, through blending other numerical methods into our IPAD framework, some previous inexact methods applied to applications \cite{Mairal2010Online,Huang2015A,Chen2016Dehazing} can also be proved to achieve the global convergence property.

\subsection{Implementable Error Bound}\label{Sec:imp}

Though we have given criteria of inexactness in the Criterion \ref{crit}, there appears another problem: the errors $\e_x^{t}$ and $\e_y^{t}$ can not be directly calculated.
Since $\mathbf{g}_x^{t}$ and $\mathbf{g}_y^{t}$ in Eq. (\ref{iPAM:iex_KKT}) are unattainable in practice, $\e_x^{t}$ and $\e_y^{t}$ can not be directly calculated by the equalities in Eq. (\ref{iPAM:iex_KKT}).
Thus the aforementioned IPAD algorithm, i.e., Alg. \ref{Alg:iPAM1} is only in the conceptual progress; the criteria (\ref{Alg_Imp:err_crit}) are not implementable in practice.

%\begin{algorithm}[!t]
%\caption{$\mathbf{u}^t = \psi(\mathbf{u}; \mathbf{u}^{t-1}, \mathbf{v}, h, H, \eta, C_u)$.}
%\begin{algorithmic}[1]
%\STATE With parameters $C_x$, $C_y$, $\{\eta_1^t\}_{t\in \mathbb{N}}$, $\{\eta_2^t\}_{t \in \mathbb{N}}$ in Alg. \ref{Alg:iPAM1}.
%\STATE Denote $\mathcal{A}$ as one step of inner iteration schemes for optimizing $h(\mathbf{u}) + H(\mathbf{u}, \mathbf{v}) + \frac{\eta}{2}\|\mathbf{u}-\mathbf{u}^{t-1}\|^2$.
%\STATE Let $\mathbf{u}^{t,0}=\mathbf{u}^{t-1}$.
%\WHILE {$\|\e_u^{t,i}\| > C_u\|\mathbf{u}^{t,i}-\mathbf{u}^{t-1}\|$}
%\STATE $\mathbf{u}^{t,i} = \mathcal{A}(\mathbf{u}^{t, i-1})$.
%\STATE $\sv_u^{t,i}=\mathbf{u}^{t,i} - \nabla_{\mathbf{u}}H(\mathbf{u}^{t,i}, \mathbf{v}) - \eta(\mathbf{u}^{t,i}-\mathbf{u}^{t-1}).$
%\STATE $\widetilde{\mathbf{u}}^{t,i} = \mathrm{prox}^{1}_{h}(\sv_u^{t,i})$.
%\STATE $\e_u^{t,i}= (1-\eta)(\widetilde{\mathbf{u}}^{t,i}-\mathbf{u}^{t,i})+\nabla_{\mathbf{u}}H(\mathbf{u}^{t,i}, \mathbf{v})-\nabla_{\mathbf{u}}H(\widetilde{\mathbf{u}}^{t,i}, \mathbf{v})$.
%\ENDWHILE
%\STATE $\mathbf{u}^{t}=\widetilde{\mathbf{u}}^{t,i}$.
%\end{algorithmic}\label{Alg:iPAM2}
%\end{algorithm}

Instead, we in this section provide a truly implementation for carrying IPAD for practical use.
After calculating the inexact solutions $\x^{t}$ and $\y^{t}$ of Eq. (\ref{iPAM:iex_sequence}) by some numerical methods, we first compute two intermediate variables $\widetilde{\x}^{t}$ and $\widetilde{\y}^{t}$ as the solutions of specific proximal mappings\footnote{$\mathrm{prox}^{\tau}_{\sigma}(\x)\in\arg\min_{\z}\sigma(\z)+\frac{\tau}{2}\|\z-\x\|^2$ denotes proximal mapping to proper, lower semi-continuous function $\sigma$.}
\begin{equation}\label{Alg_Imp:extra_variables}
\widetilde{\x}^{t}=\mathrm{prox}^{1}_{f}(\sv_x^{t}), \ \ \ \  \widetilde{\y}^{t}=\mathrm{prox}^{1}_{g}(\sv_y^{t}),
\end{equation}
where variables $\sv_x^t$ and $\sv_y^t$ are computed as follows:
\begin{equation}\label{eq:xx}
\begin{aligned}
\sv_x^{t}&=\x^{t} - \nabla_{\x}H(\x^{t}, \y^{t-1}) - \eta_1^{t-1}(\x^{t}-\x^{t-1}),\\
\sv_y^{t}&=\y^{t} - \nabla_{\y}H(\x^{t}, \y^{t}) - \eta_2^{t-1}(\y^{t}-\y^{t-1}).
\end{aligned}
\end{equation}
After getting the intermediate variables $\widetilde{\x}^{t}$ and $\widetilde{\y}^{t}$ from $\x^{t}$ and $\y^{t}$, the errors $\e_x^t$ and $\e_y^t$ are calculated by:
\begin{equation}\label{Alg_Imp:implem_err}\small
\begin{aligned}
\e_x^{t}&= (1-\eta_1^{t-1})(\widetilde{\x}^{t}-\x^{t})+\nabla_{\x}H(\x^{t}, \y^{t-1})-\nabla_{\x}H(\widetilde{\x}^{t}, \y^{t-1}),\\
\e_y^{t}&=(1-\eta_2^{t-1})(\widetilde{\y}^{t}-\y^{t})+\nabla_{\y}H(\x^{t}, \y^{t})-\nabla_{\y}H(\x^{t}, \widetilde{\y}^{t}).\\
\end{aligned}
\end{equation}

We give a proposition as follows to show that these two errors are exactly the implementable versions of Eq. (\ref{iPAM:iex_KKT}).
In order to make discussions smoother, we provide the proof of Proposition \ref{prop} at the end of this paper in Appendix A.
\begin{prop}\label{prop}
The errors, $\e_x^t$ and $\e_y^t$ calculated though the Eq. (\ref{Alg_Imp:implem_err}) are equivalent to the ones in Eq. (\ref{iPAM:iex_KKT}).
\end{prop}

Thus, the errors in (\ref{Alg_Imp:implem_err}) are used for checking whether the stopping criteria (\ref{Alg_Imp:err_crit}) are satisfied; once the criteria are satisfied, we re-assign $\widetilde{\x}^{t}$ to $\x^{t}$ and $\widetilde{\y}^{t}$ to $\y^{t}$.
Then $\widetilde{\x}^{t}$ and $\widetilde{\y}^{t}$ become the final solutions of their corresponding subproblems at $t$-th step.
For clarity, we give the detailed inexact process of inner iteration schemes in Alg. \ref{Alg:iPAM2}.

%%%%%%%%%%%%%%%%%%%%%%%%%%%%%%%%%%%%%%%%%%%%%%%%%%%%%%%%%%%%%%%%%%%%%%%%%%%%%%%%%%%%%%%%%%%%%%%%%%%%%%%%%%%%%%%%%%%%%%%%%%%%%%%%%%%%%%%%%%%%%%%%%%%%%%%%%%%%%%%%%%
%%%%%%%%%%%%%%%%%%%%%%%%%%%%%%%%%%%%%%%%%%%%%%%%%%%%%%%%%%%%%%%%%%%%%%%%%%%%%%%%%%%%%%%%%%%%%%%%%%%%%%%%%%%%%%%%%%%%%%%%%%%%%%%%%%%%%%%%%%%%%%%%%%%%%%%%%%%%%%%%%%%%%%%%

\begin{figure*}[!t]
\setlength{\abovecaptionskip}{0cm} %缩小caption和图像之间的距离
\setlength{\belowcaptionskip}{0cm} %缩小caption和下方文字的距离
%\hspace{0.2cm}
\begin{minipage}[b]{0.245\linewidth}
  \centering
  \centerline{\includegraphics[width=1.1\textwidth]{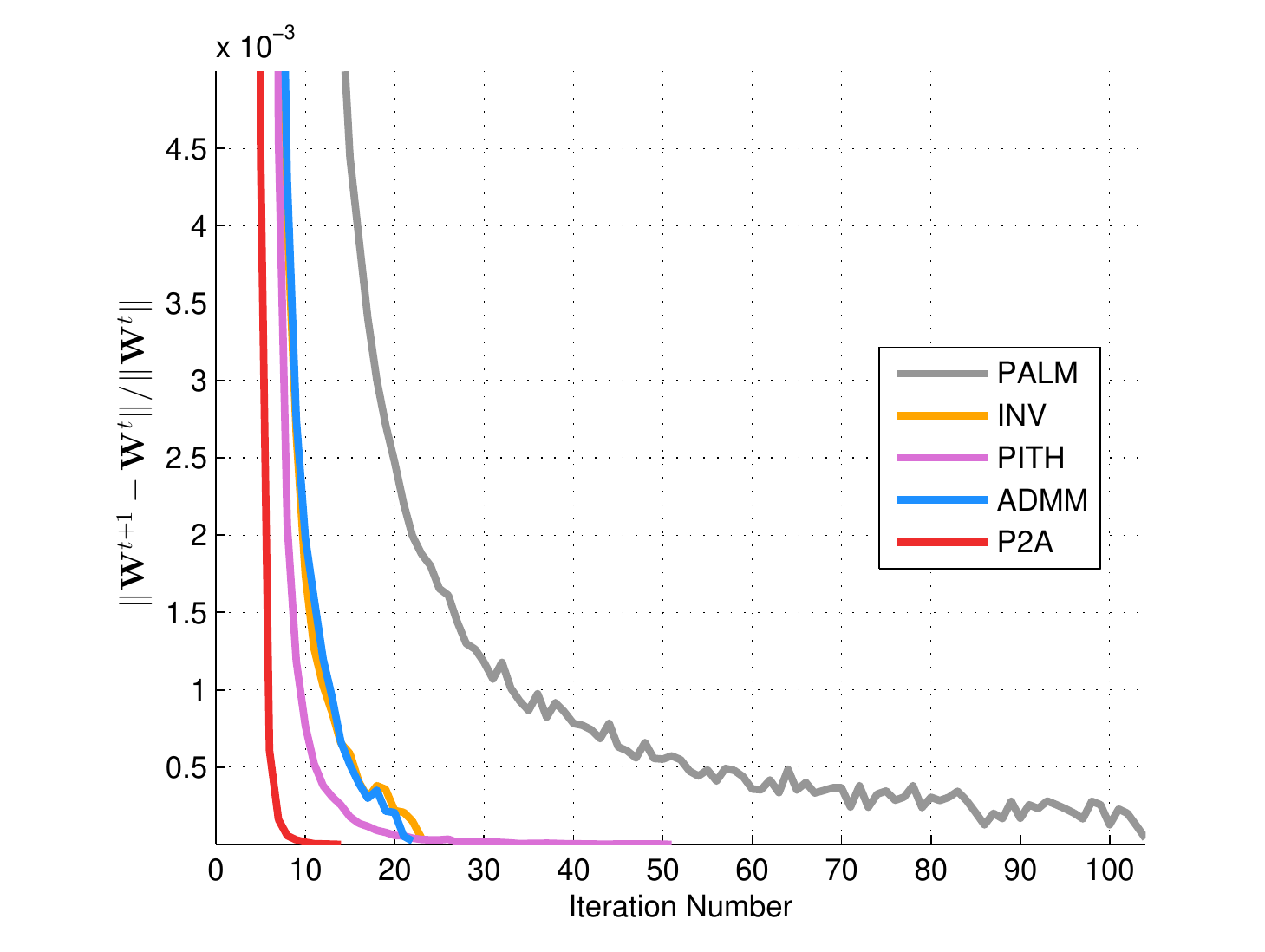}}
 %\vspace{1.5cm}
  \centerline{\quad \ \small(a) \begin{scriptsize}$\|\W^{t+1}-\W^t\|/\|\W^t\|$\end{scriptsize}}\medskip
   \vspace{-0.6cm}
\end{minipage}
\hspace{-0.2cm}
\hfill
\begin{minipage}[b]{0.245\linewidth}
  \centering
  \centerline{\includegraphics[width=1.1\textwidth]{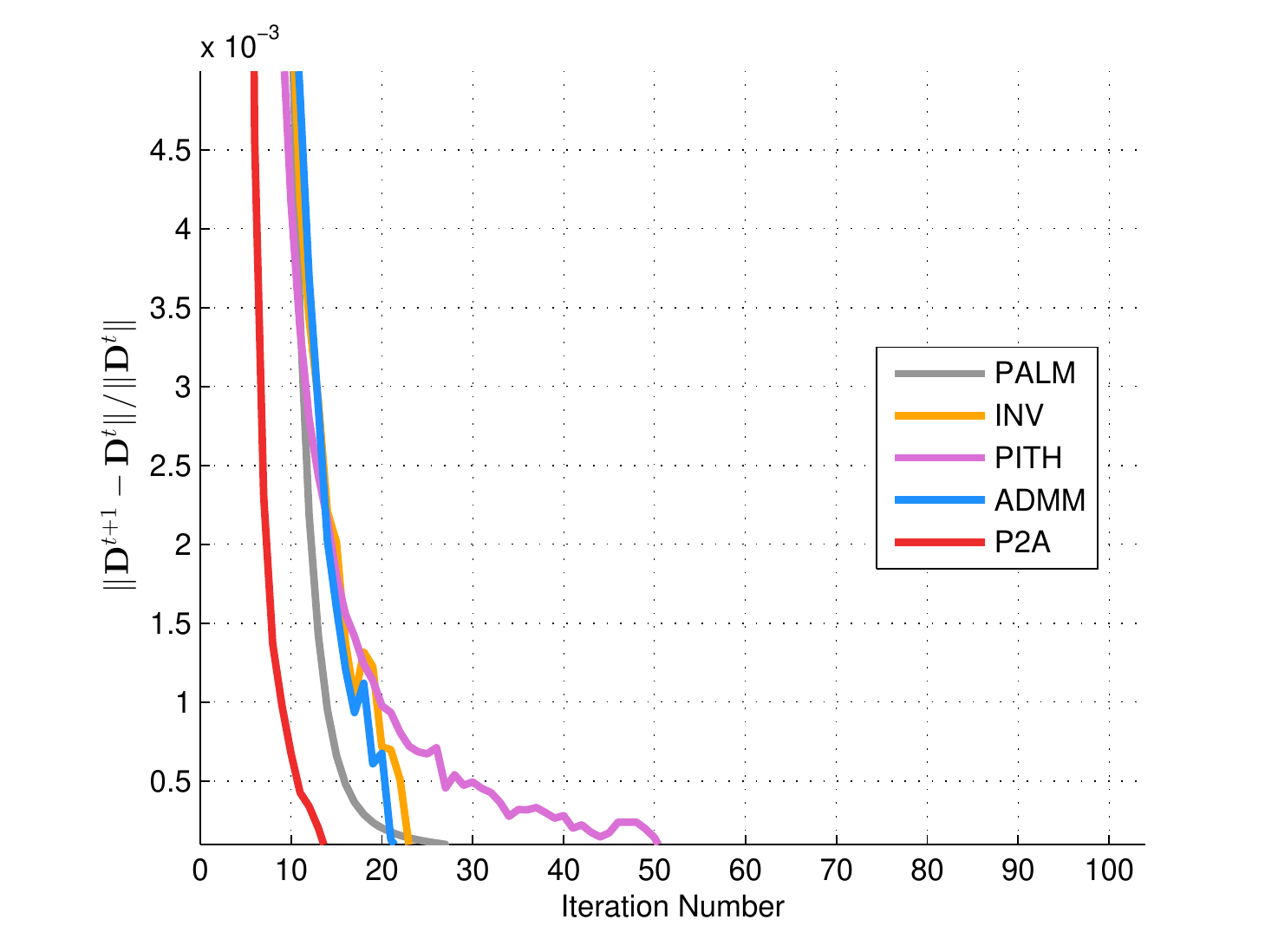}}
 %\vspace{1.5cm}
  \centerline{\quad \ \small (b) \begin{scriptsize}$\|\D^{t+1}-\D^t\|/\|\D^t\|$\end{scriptsize}}\medskip
   \vspace{-0.6cm}
\end{minipage}
\hspace{-0.2cm}
\hfill
\begin{minipage}[b]{0.245\linewidth}
  \centering
  \centerline{\includegraphics[width=1.1\textwidth]{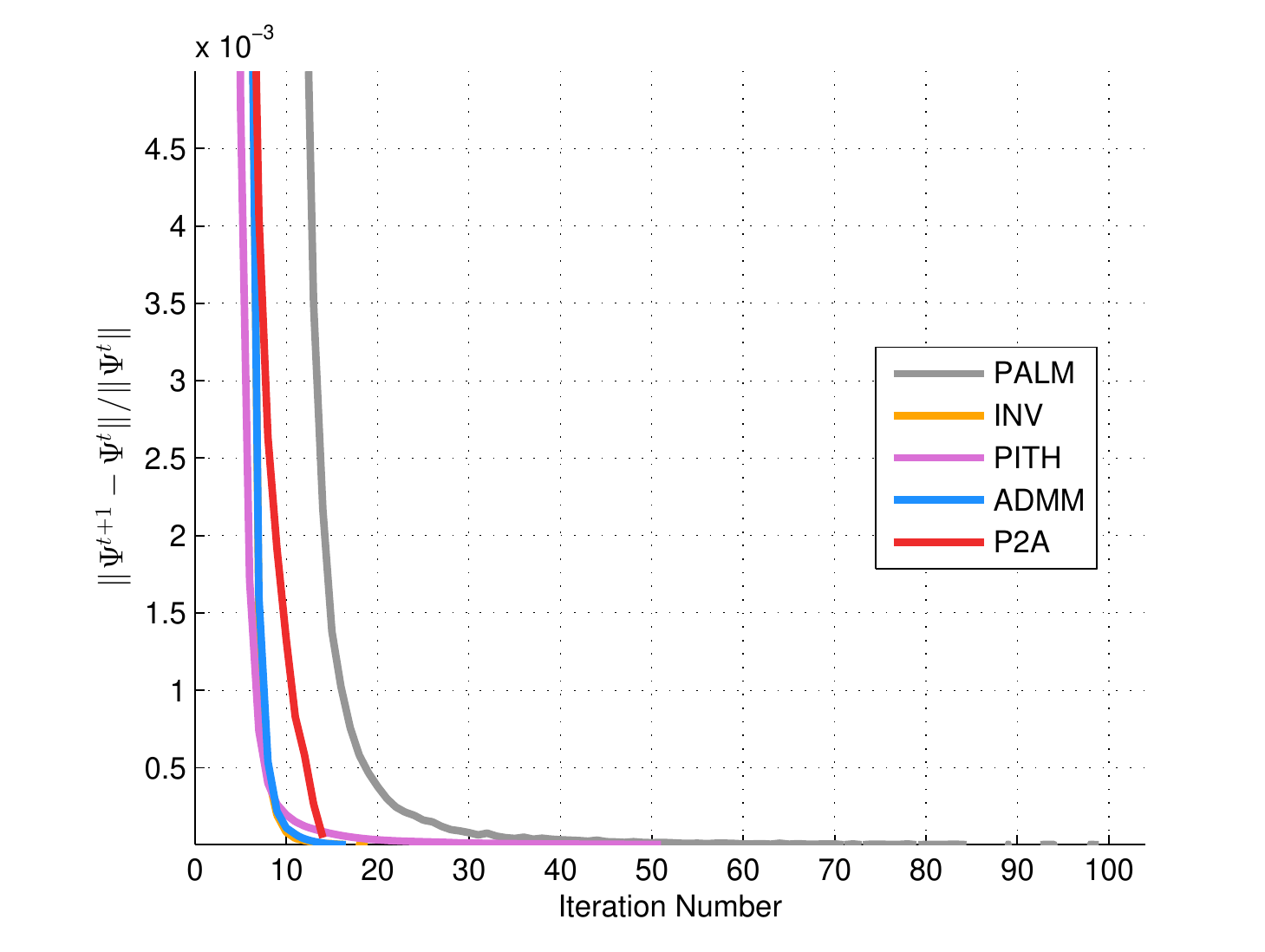}}
 %\vspace{1.5cm}
  \centerline{\quad \ \small (c) \begin{scriptsize}$\|\Psi^{t+1}-\Psi^t\|/\|\Psi^t\|$\end{scriptsize}}\medskip
   \vspace{-0.6cm}
\end{minipage}
\hspace{-0.2cm}
\hfill
\begin{minipage}[b]{0.245\linewidth}
  \centering
  \centerline{\includegraphics[width=1.1\textwidth]{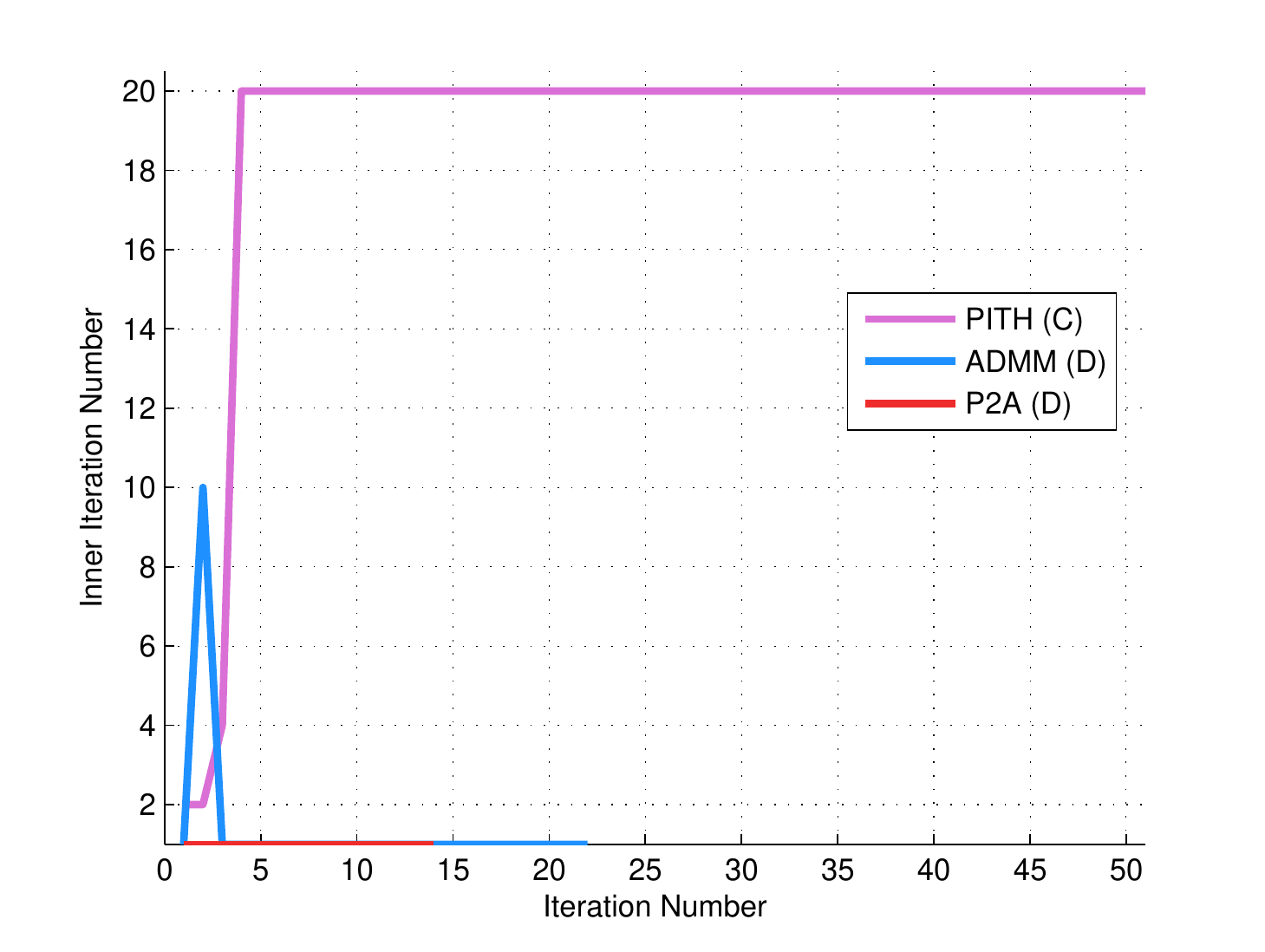}}
 %\vspace{1.5cm}
  \centerline{\quad \ \small (d) \begin{scriptsize}Inner Iteration\end{scriptsize}}\medskip
   \vspace{-0.6cm}
\end{minipage}
\hspace{0.2cm}
\end{figure*}

\begin{figure*}[!t]
\setlength{\abovecaptionskip}{0cm} %缩小caption和图像之间的距离
\setlength{\belowcaptionskip}{0cm} %缩小caption和下方文字的距离
%\hspace{0.2cm}
\begin{minipage}[b]{0.245\linewidth}
  \centering
  \centerline{\includegraphics[width=1.1\textwidth]{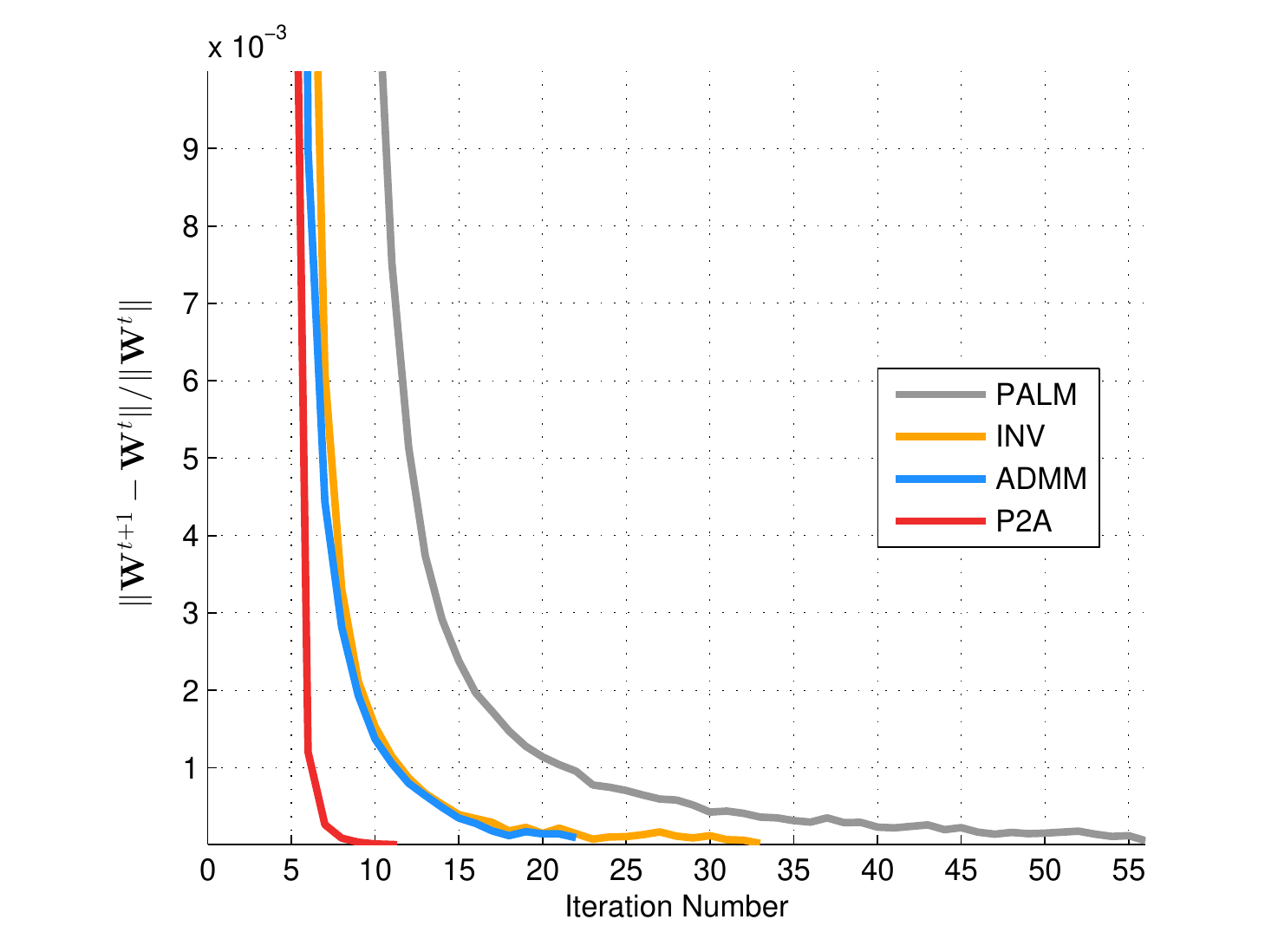}}
 %\vspace{1.5cm}
  \centerline{\quad \ \small(e) \begin{scriptsize}$\|\W^{t+1}-\W^t\|/\|\W^t\|$\end{scriptsize}}\medskip
   \vspace{-0.6cm}
\end{minipage}
\hspace{-0.2cm}
\hfill
\begin{minipage}[b]{0.245\linewidth}
  \centering
  \centerline{\includegraphics[width=1.1\textwidth]{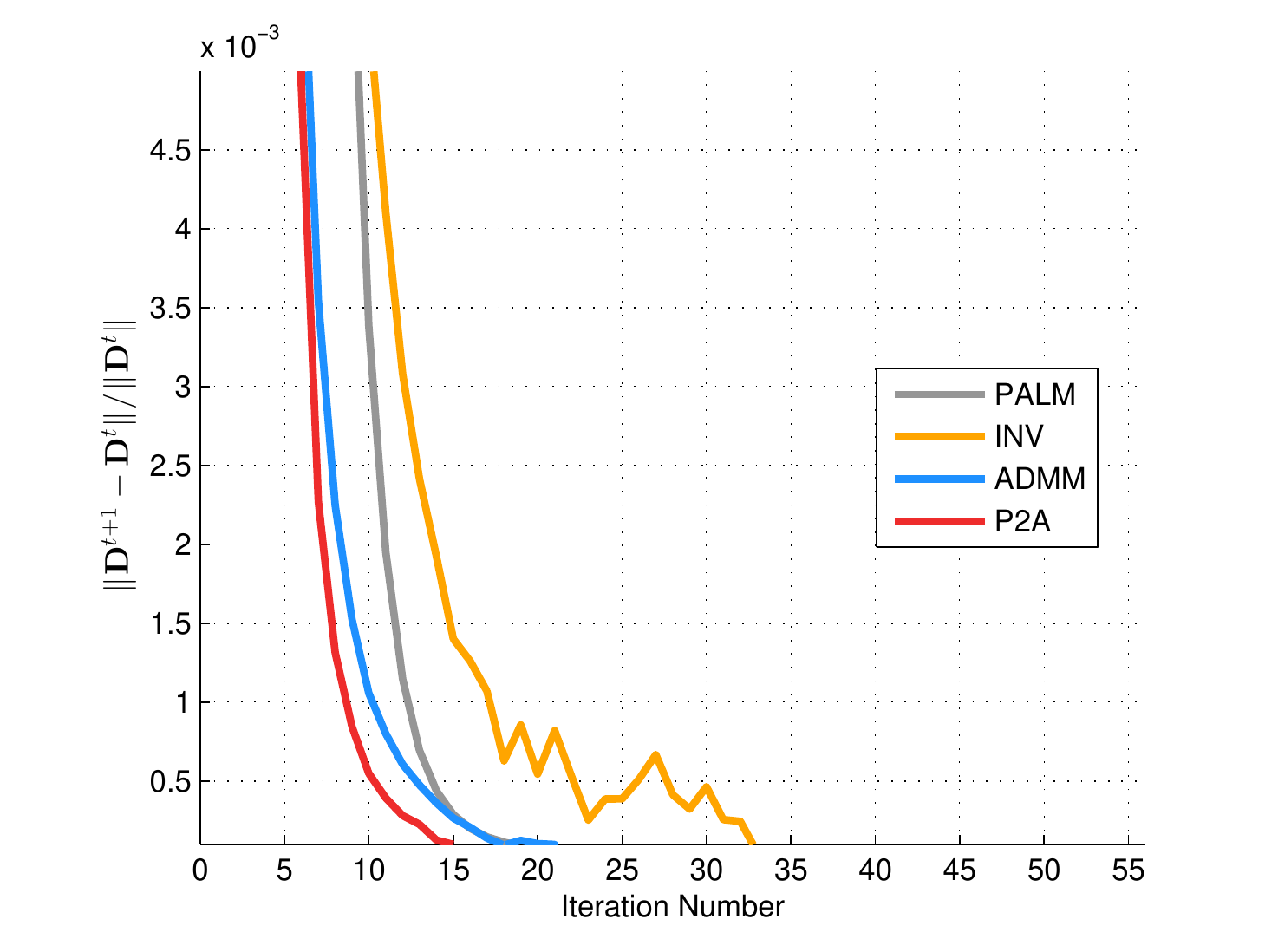}}
 %\vspace{1.5cm}
  \centerline{\quad \ \small (f) \begin{scriptsize}$\|\D^{t+1}-\D^t\|/\|\D^t\|$\end{scriptsize}}\medskip
   \vspace{-0.6cm}
\end{minipage}
\hspace{-0.2cm}
\hfill
\begin{minipage}[b]{0.245\linewidth}
  \centering
  \centerline{\includegraphics[width=1.1\textwidth]{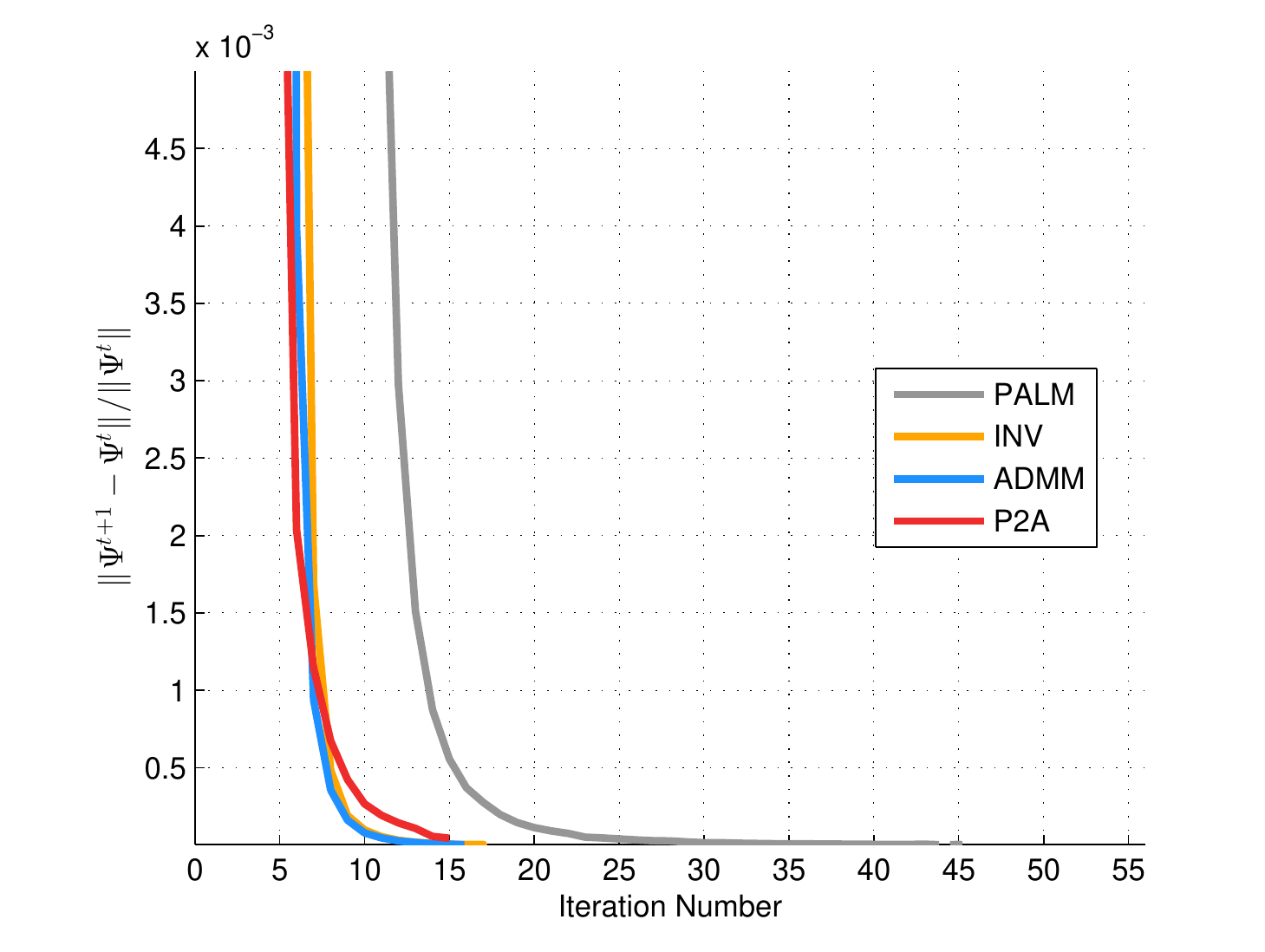}}
 %\vspace{1.5cm}
  \centerline{\quad \ \small (g) \begin{scriptsize}$\|\Psi^{t+1}-\Psi^t\|/\|\Psi^t\|$\end{scriptsize}}\medskip
   \vspace{-0.6cm}
\end{minipage}
\hspace{-0.2cm}
\hfill
\begin{minipage}[b]{0.245\linewidth}
  \centering
  \centerline{\includegraphics[width=1.1\textwidth]{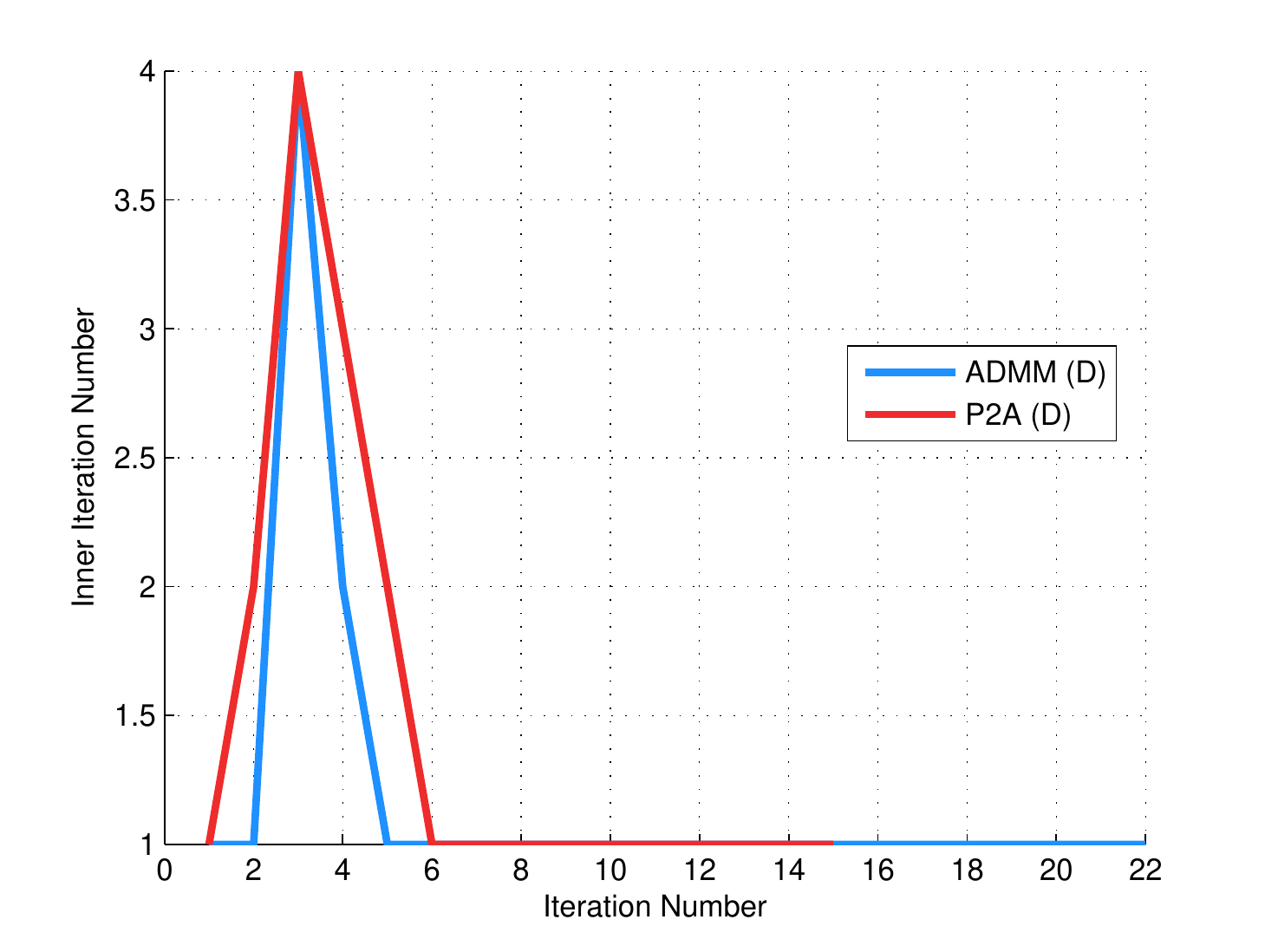}}
 %\vspace{1.5cm}
  \centerline{\quad \ \small (h) \begin{scriptsize}Inner Iteration\end{scriptsize}}\medskip
   \vspace{-0.6cm}
\end{minipage}
\hspace{0.2cm}
\end{figure*}

\begin{figure*}[!t]
\setlength{\abovecaptionskip}{0cm} %缩小caption和图像之间的距离
\setlength{\belowcaptionskip}{0cm} %缩小caption和下方文字的距离
%\hspace{0.2cm}
\begin{minipage}[b]{0.245\linewidth}
  \centering
  \centerline{\includegraphics[width=1.1\textwidth]{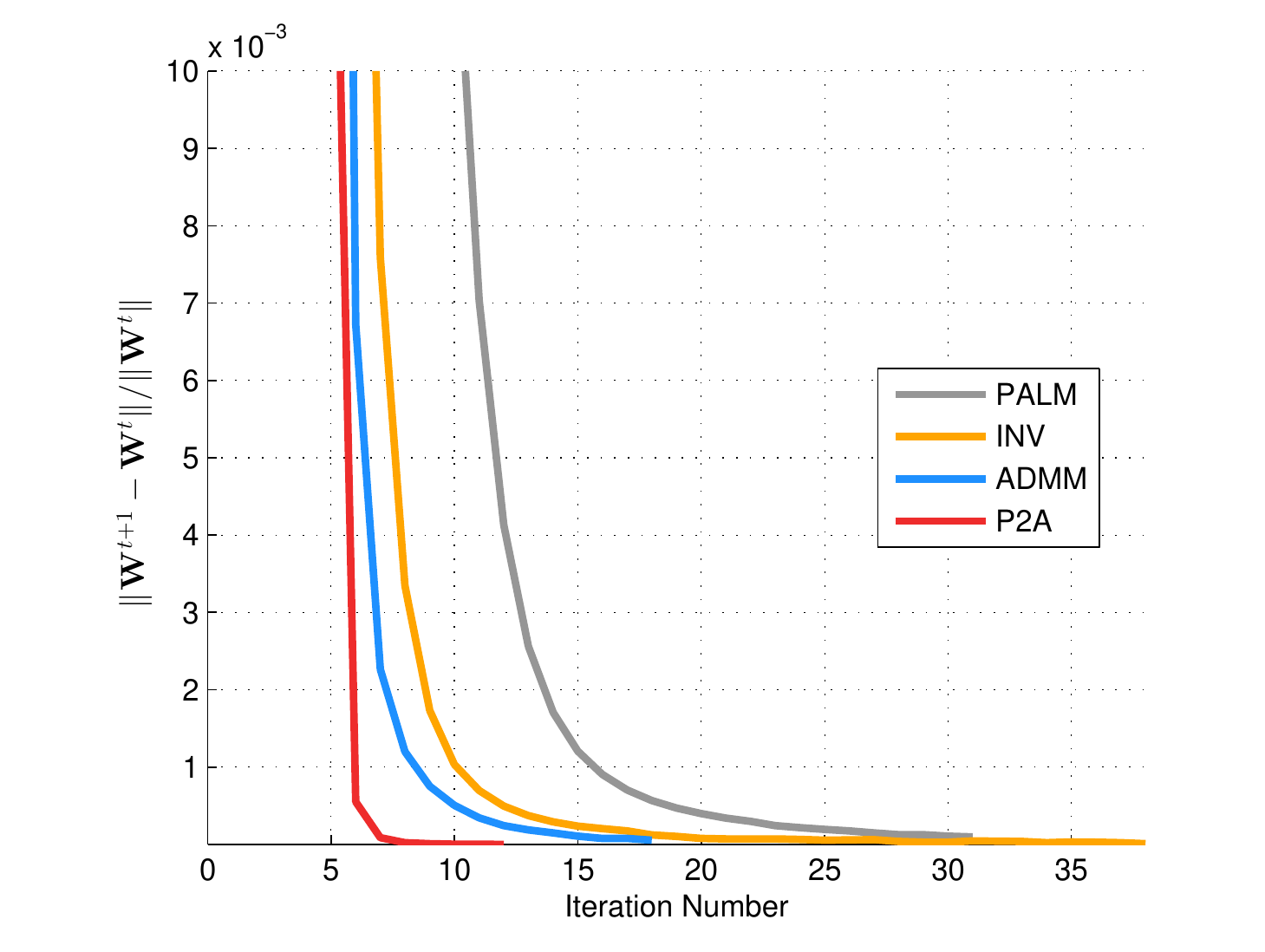}}
 %\vspace{1.5cm}
  \centerline{\quad \ \small(i) \begin{scriptsize}$\|\W^{t+1}-\W^t\|/\|\W^t\|$\end{scriptsize}}\medskip
   \vspace{0.2cm}
\end{minipage}
\hspace{-0.2cm}
\hfill
\begin{minipage}[b]{0.245\linewidth}
  \centering
  \centerline{\includegraphics[width=1.1\textwidth]{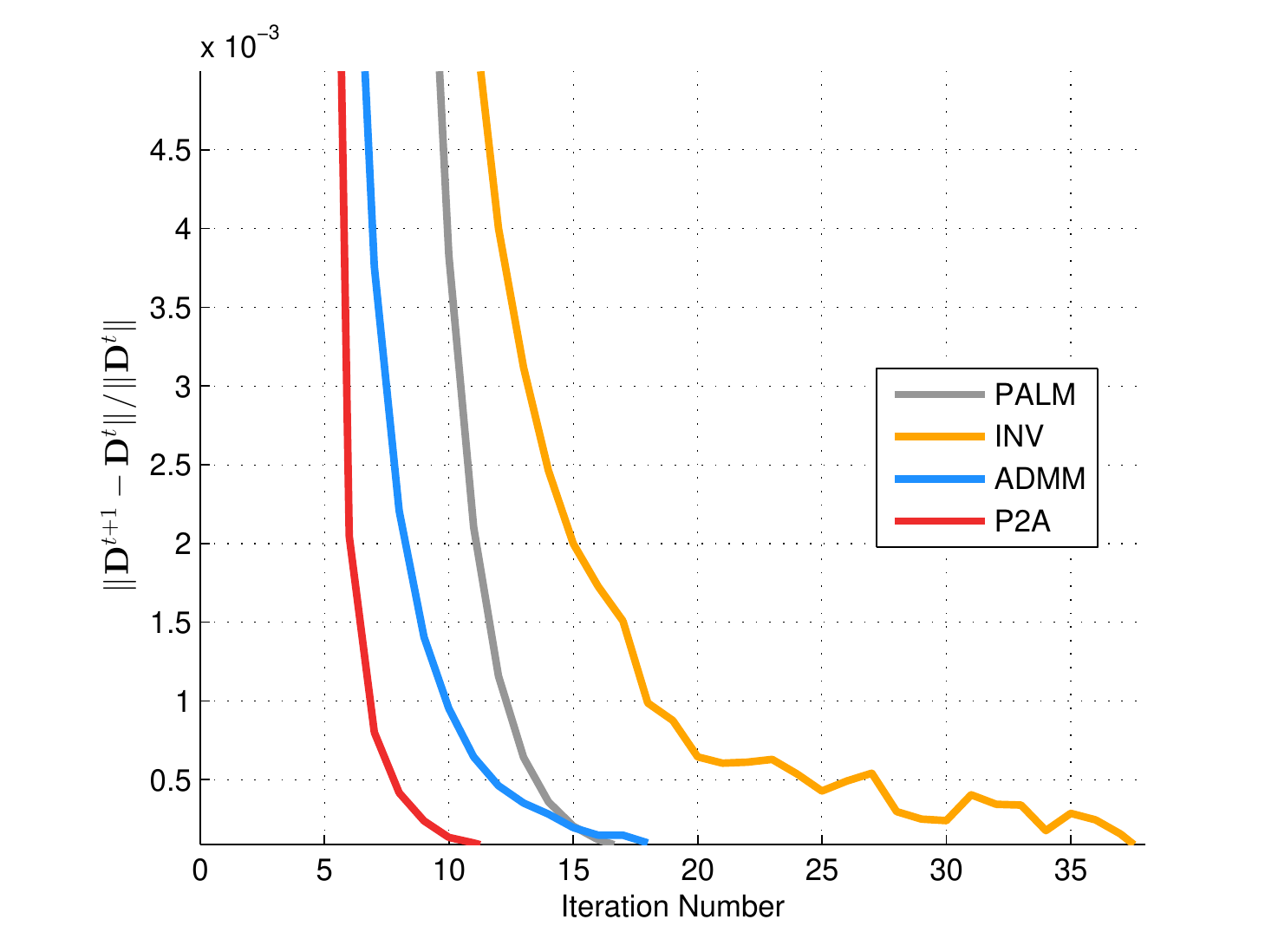}}
 %\vspace{1.5cm}
  \centerline{\quad \ \small (j) \begin{scriptsize}$\|\D^{t+1}-\D^t\|/\|\D^t\|$\end{scriptsize}}\medskip
   \vspace{0.2cm}
\end{minipage}
\hspace{-0.2cm}
\hfill
\begin{minipage}[b]{0.245\linewidth}
  \centering
  \centerline{\includegraphics[width=1.1\textwidth]{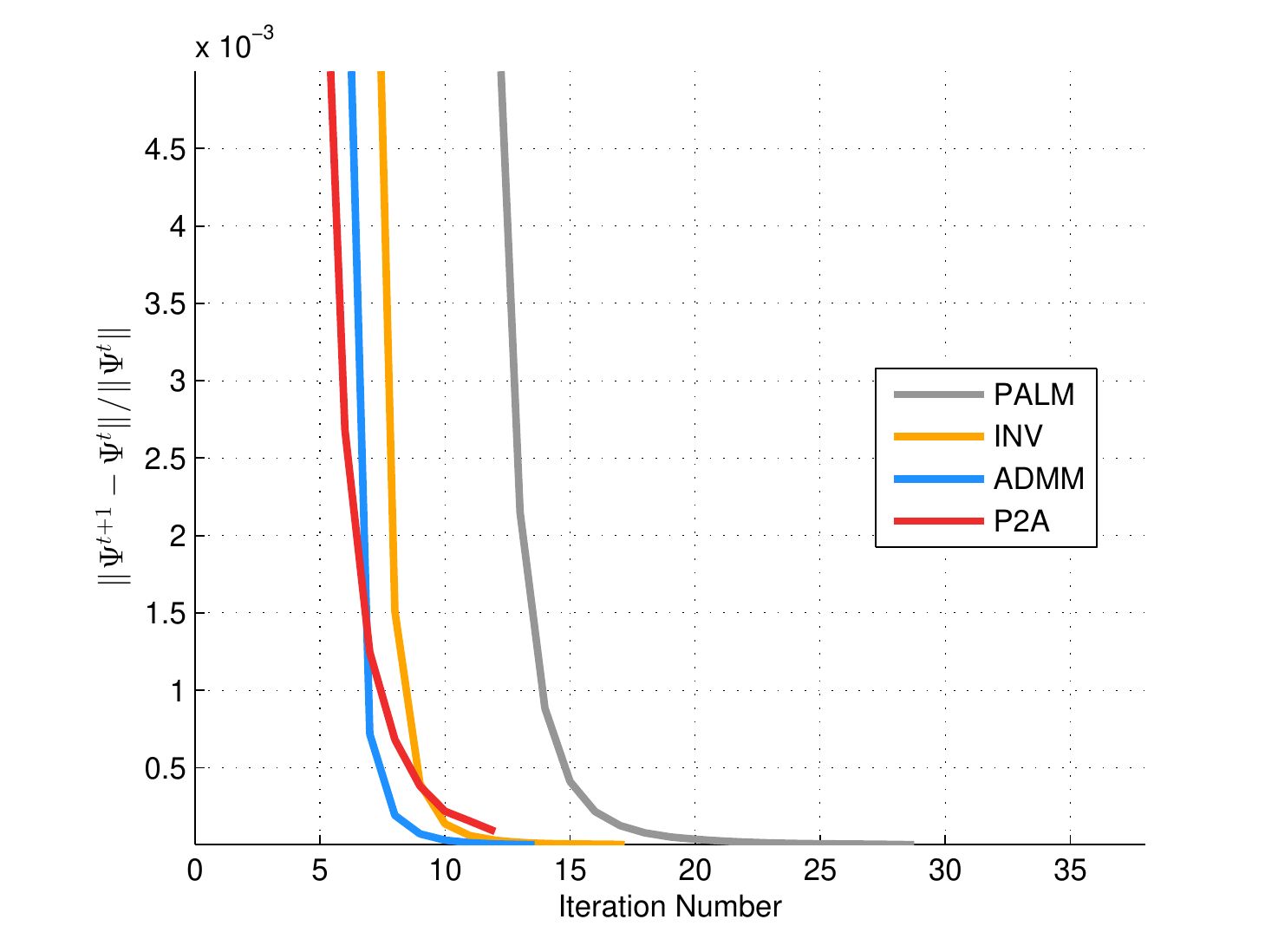}}
 %\vspace{1.5cm}
  \centerline{\quad \ \small (k) \begin{scriptsize}$\|\Psi^{t+1}-\Psi^t\|/\|\Psi^t\|$\end{scriptsize}}\medskip
   \vspace{0.2cm}
\end{minipage}
\hspace{-0.2cm}
\hfill
\begin{minipage}[b]{0.245\linewidth}
  \centering
  \centerline{\includegraphics[width=1.1\textwidth]{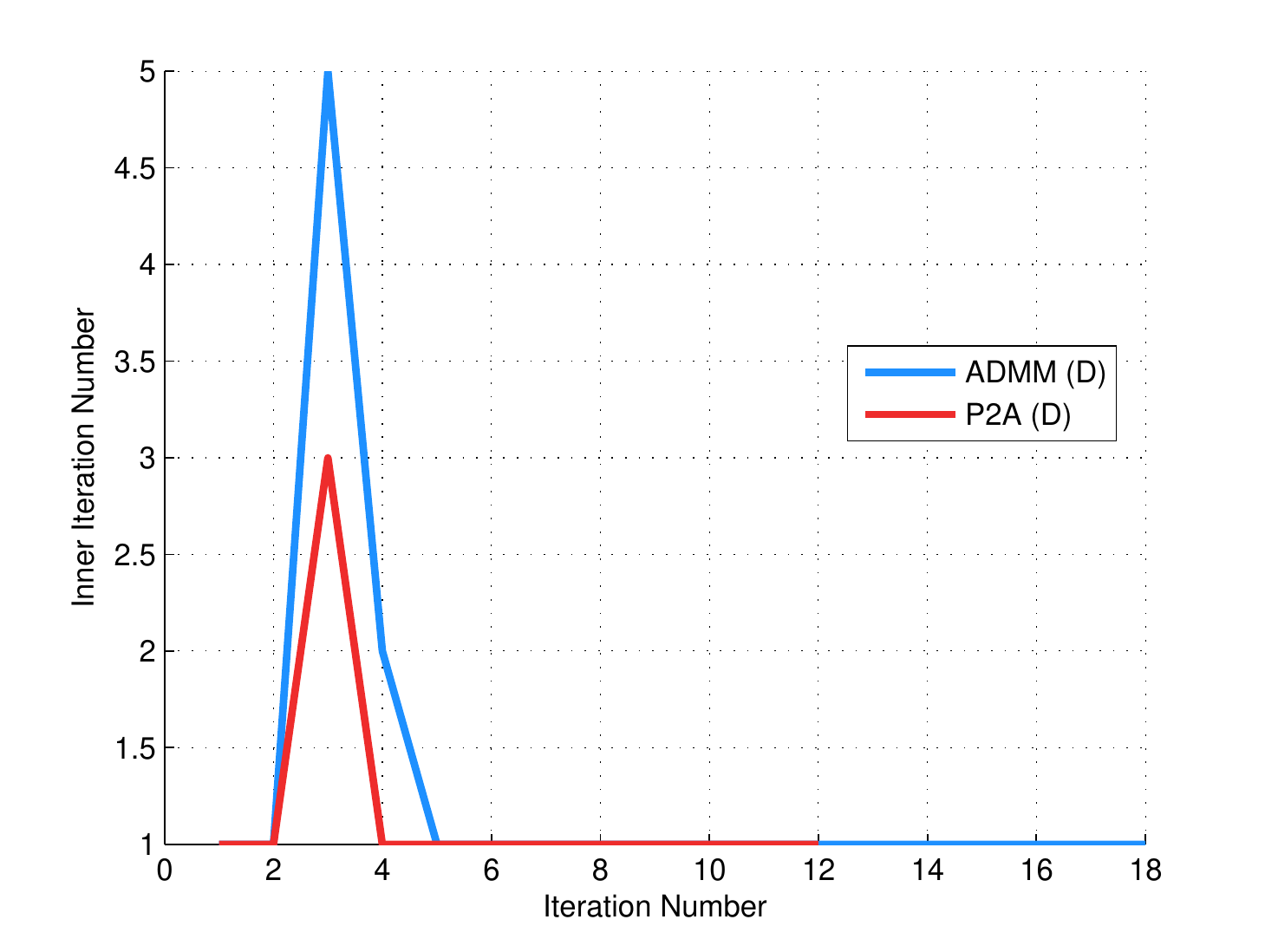}}
 %\vspace{1.5cm}
  \centerline{\quad \ \small (l) \begin{scriptsize}Inner Iteration\end{scriptsize}}\medskip
  \vspace{0.2cm}
\end{minipage}
\hspace{0.2cm}

\caption{The convergence properties of using PALM \cite{bolte2014proximal}, INV \cite{bao2016dictionary}, IPAD-PITH (PITH for short in the figures), IPAD-ADMM (ADMM for short) and IPAD-P2A (P2A for short) for SDL problem with $\ell_0$ penalty on synthetic data.
The convergence results in the first row belongs to $n=64$; the second row belongs to $n=144$ and the last row belongs to $n=256$.
%(\textbf{Best viewed on screen!})
}
\label{Fig:converge_synthetic}
\end{figure*}

\section{Experimental Results}\label{Sec:Experiments}
Though our IPAD algorithm framework can be directly applied to numerous problems in computer vision and machine learning \cite{Zou2012Sparse,Huang2015A,Shi2011A,Chen2016Dehazing},
we in this paper just consider to utilize the widely concerned $\ell_0$-regularized SDL model \cite{bao2016dictionary} as an example to verify the flexibility and efficiency of our IPAD framework.
All the compared algorithms are implemented by Matlab R2013b and tested on a PC with 8 GB of RAM and Intel Core i5-4200M CPU.

\subsection{SDL with $\ell_0$ Penalty}
The SDL problem with $\ell_0$ penalty is formulated as:
\begin{equation}\label{Exp:DL_model}
\min_{\D, \W} \frac{1}{2}\|\I - \D\W^{\top}\|^2 + \lambda\|\W\|_0  +\mathcal{X}_{\mathcal{W}}(\W) + \mathcal{X}_{\mathcal{D}}(\D),
\end{equation}
where $\|\cdot\|_0$ denotes the $\ell_0$ penalty that counts the number of non-zero elements of $\W$.
Here the indicator function $\mathcal{X}$ acts on the set
$$\mathcal{D}=\{\D=\{\bd_i\}_{i=1}^m \subset \mathbb{R}^{n\times m}: \|\bd_i\|=1, \forall i\}.$$
As for another set $\mathcal{W}$,
we make it empty for synthetic data but define it as
$$\mathcal{W}=\{\W=\{\w_i\}_{i=1}^m\subset \mathbb{R}^{p\times m}: \|\w_i\|_{\infty}\leq U_b, \forall i\},$$
for real-world data to enhance the stability of the model \cite{bao2016dictionary}.
Moreover, we denote
$\mathcal{S}(\W)=\lambda\|\W\|_0 +\mathcal{X}_{\mathcal{W}}(\W)$
to simplify the deduction.

It is observed that problem (\ref{Exp:DL_model}) is a special case of problem (\ref{eq:2variable}).
Thus, we can apply IPAD for inexactly solving the following subproblems since it is extremely hard to get exact solutions of the subproblems.
\begin{equation}\label{Exp:inexact_W}\small
\min_{\W} \mathcal{S}(\W) + \frac{1}{2}\|\I - \D^{t}\W^{\top}\|^2 + \frac{\eta_1^{t}}{2}\|\W-\W^t\|^2,
\end{equation}
\begin{equation}\label{Exp:inexact_D}\small
\min_{\D} \mathcal{X}_{\mathcal{D}}(\D) + \frac{1}{2}\|\I - \D(\W^{t+1})^{\top}\|^2 + \frac{\eta_2^{t}}{2}\|\D-\D^t\|^2.
\end{equation}
Moreover, PALM can also be applied to solve (\ref{Exp:DL_model}), but it requires computing Lipschitz constants at every iteration.

Notice that solving the nonconvex subproblem (\ref{Exp:inexact_W}) is not a trivial task.
Here we apply a proximal iterative hard-thresholding (PITH) algorithm \cite{Bach2011Optimization,Herrity2006Sparse} to solve this subproblem.
On the other hand, we apply ADMM \cite{boyd2011distributed} for solving subproblem (\ref{Exp:inexact_D}).

\subsection{Synthetic Data}
We generate synthetic data with different sizes to help analyze the property of IPAD (see Table \ref{Tab:Synthetic}).
All the algorithms for the synthetic data stop when satisfying:
\begin{equation}\footnotesize
\max\left\{\frac{\|\D^{t+1}-\D^{t}\|}{\|\D^t\|}, \ \frac{\|\W^{t+1}-\W^{t}\|}{\|\W^t\|}, \  \frac{\|\Psi^{t+1}-\Psi^{t}\|}{\|\Psi^t\|}\right\}<1e^{-4},
\end{equation}
where $\Psi^{t}$ is the objective value at step $t$.

\begin{table*}[!t]\small
\begin{center}
\begin{tabular}{|c|c|c|c|c|c|c|c|c|c|c|c|c|c|}\hline
Data & \multicolumn{5}{c|}{$n=64, m= 600, p=4000$} & \multicolumn{4}{c|}{$n=144, m= 900, p=10000$} &  \multicolumn{4}{c|}{$n=256, m= 1600, p=16000$}\\ \hline
Alg. & PALM  &   INV &   PITH  &   ADMM & P2A   &  PALM & INV   & ADMM  &  P2A  &  PALM &  INV &  ADMM & P2A \\
\hline
Out-iter & 104   &  23  &  51    &  22   &  \textbf{14}   &  56  & 33    & 22     &  \textbf{15}  &  31 & 38 &   18 & \textbf{12}\\
\hline
Time (s) & 52.82 & \textbf{7.81} & 253.56 &  8.72 &  8.31 & 96.08 & 45.93 & \textbf{35.58} & 38.94 & 319.97 &   253.17 & \textbf{150.25} & 158.42\\
\hline
 \end{tabular}
\end{center}
\caption{The number of outer iterations and the iteration time (s) of PALM \cite{bolte2014proximal}, INV \cite{bao2016dictionary}, our IPAD-PITH (PITH for short in this table), our IPAD-ADMM (ADMM for short) and our IPAD-P2A (P2A for short) for synthetic data.
	The results are the averages of multiple tests.}\label{Tab:Synthetic}
\end{table*}

\subsubsection{Efficiency of Inexact Strategy} \label{sec:4-2-1}

To show the respective effects of using inexact strategies on different subproblems, we propose IPAD-PITH which obtains $\W^{t+1}$ by PITH but keeps $\D$-subproblem the same as PALM\footnote{Due to the space limit, we give detailed algorithm implementations, further theoretical analyses and more results in supplemental material.}.
We also design IPAD-ADMM that computes $\D^{t+1}$ by ADMM but remains $\W$-subproblem the same as PALM.

\begin{table*}[!t]\small
\begin{center}
\begin{tabular}{|c||c|c|c||c|c|c||c|c|c||c|c|c|}\hline
Image / $\sigma$ / $\lambda$                    &  \multicolumn{3}{c||}{PALM \cite{bolte2014proximal}} &  \multicolumn{3}{c||}{mPALM \cite{bao2014l0}}  &   \multicolumn{3}{c||}{INV \cite{bao2016dictionary}} &  \multicolumn{3}{c|}{IPAD-ADMM}      \\ \hline
\multirow{2}{*}{``Peppers512'' / $30$ / $5500$}
                                &  PSNR & Iter & Time        &  PSNR & Iter & Time        &  PSNR & Iter & Time        &  PSNR & Iter & Time      \\ \cline{2-13}
                                &28.64  & 4 & 4.11           &30.11 &10 & 325.53          &30.14 & 61& 48.63           &\textbf{30.21}  & 25 & 23.29            \\ \hline
\multirow{2}{*}{``Lena512'' / $25$ / $4500$}
                                &  PSNR & Iter & Time        &  PSNR & Iter & Time        &  PSNR & Iter & Time        &  PSNR & Iter & Time      \\ \cline{2-13}
                                &28.85  & 4 & 3.98           &31.04  &14 & 459.10          &31.11 & 62& 50.20           &\textbf{31.13}  & 31 & 37.78            \\ \hline
\multirow{2}{*}{``Barbara512'' / $20$ / $3500$}
                                &  PSNR & Iter & Time        &  PSNR & Iter & Time        &  PSNR & Iter & Time        &  PSNR & Iter & Time     \\ \cline{2-13}
                                &28.73  & 4 & 4.13          &30.06  & 11 & 399.19        &30.09  & 58 & 45.09         &\textbf{30.22}  & 18 & 16.54            \\ \hline
\multirow{2}{3cm}{``Hill512'' / $15$ / $2500$}
                                &  PSNR &Iter & Time         &  PSNR & Iter & Time        &  PSNR & Iter & Time        &  PSNR & Iter & Time      \\ \cline{2-13}
                                &29.84 & 4 & 4.03           &31.20  &26 & 85.34         &31.31  & 84 & 69.73         &\textbf{31.32}  & 19 &17.81            \\ \hline
 \end{tabular}
\end{center}
\caption{The PSNR scores of the recovered images, number of outer iterations and the whole iteration time (s) of PALM \cite{bolte2014proximal}, mPALM \cite{bao2014l0}, INV \cite{bao2016dictionary} and our IPAD-ADMM for real-world data.}\label{Tab:Real-world}
\end{table*}

\begin{table*}[!t]\small
	\begin{center}
		\begin{tabular}{|l|c|c|c|c|}\hline
			Method                     &  $\sigma = 30$ & $\sigma = 25$ & $\sigma = 20$ & $\sigma = 15$    \\ \hline
			\multirow{2}{0.4cm}{KSVD \cite{elad2006image}}
			&  PSNR / Time         & PSNR / Time        & PSNR / Time        &  PSNR / Time       \\ \cline{2-5}
			&  \textbf{29.21} / 184.88      & \textbf{30.09} / 235.22     & \textbf{31.14} / 318.06     &  \textbf{32.49} / 481.48     \\ \hline
			\multirow{2}{0.4cm}{IPAD-ADMM}
			&  PSNR / Time         & PSNR / Time        & PSNR / Time        &  PSNR / Time     \\ \cline{2-5}
			& 28.98 / \textbf{20.04}       & 29.85 / \textbf{26.75}      & 30.89 / \textbf{22.72}      & 32.28 / \textbf{22.65}     \\ \hline
		\end{tabular}
	\end{center}
	\caption{Quantitative denoising results of KSVD \cite{elad2006image} and IPAD-ADMM on 7 widely-used example images.}\label{Tab:KSVD}
\end{table*}

The comparisons in Table \ref{Tab:Synthetic} among PALM, IPAD-PITH and IPAD-ADMM show that inexact strategies help reduce the iteration steps:
both the IPAD-PITH and IPAD-ADMM converges with less iterations than PALM.
However, the performances of IPAD-PITH and IPAD-ADMM are quite different in inner iterations.
We can see from Figure \ref{Fig:converge_synthetic}(d) that IPAD-ADMM uses few inner steps during iterations.
However, IPAD-PITH reaches the maximum inner steps (set as 20) at almost every iteration.
This from one side shows that ADMM is suitable for solving (\ref{Exp:inexact_D}) but PITH is less efficient for solving (\ref{Exp:DL_model}).
On the other side it is caused by the problems themselves: (\ref{Exp:inexact_D}) has unique solution but problem (\ref{Exp:inexact_W}) is a challenging NP hard problem and only sub-optimal solution can be found in polynomial time.
Since PITH converges with unexpected time, thus we only test it on the data with relatively low dimension ($n=64$).

We also apply inexact strategies to both subproblems of $\D$ and $\W$.
Since PITH is time-consuming for solving (\ref{Exp:inexact_W}), thus we reduce the maximum inner step to 2 for PITH.
We name this algorithm as IPAD-P2A.
Then we can see from the Figure \ref{Fig:converge_synthetic} and Table \ref{Tab:Synthetic} that IPAD-P2A uses less iteration steps and sometimes converges faster to an optimal solution.

%We can see from Fig. \ref{Fig:converge_synthetic}(a)-(c) that both iPAM-PITH and iPAM-ADMM are converged, which is consistent with the assertion in Remark \ref{Exp:Syn:remark1}.
%In addition, t

\begin{remark}\label{Exp:Syn:remark1}
Though Theorem \ref{Converg_Ana:main_theorem} is proved for applying the inexact strategy to both subproblems, the two hybrid forms of IPAD, i.e. IPAD-PITH and IPAD-ADMM, are also converged.
Furthermore, a hybrid IPAD optionally combines PALM, PAM and IPAD can be proved to have global convergence property for solving problem (\ref{eq:2variable})
\end{remark}
\begin{remark} It can be observed that IPAD-P2A can not be seen as a special case of the hybrid IPAD since it contains one inexact strategy for $\D$ and two-steps prox-linear iterations for $\W$.
But fortunately, the experimental results verify the convergence of IPAD-P2A and we can also prove the convergence from theoretical analyses.
\end{remark}

By comparing the algorithms, all the inexact strategies of our IPAD method perform better than PALM and are verified to be practicable, converged and efficient.
However, a less efficient numerical method for solving subproblem indeed reduce the efficiency of the whole algorithm.
Therefore, we suggest carefully choosing effective numerical methods for solving subproblems.

\subsubsection{Other Comparisons}
At last, we compare INV that solves $\W^{t+1}$ in the same way as PALM but treats $\D^{t+1}$ as the solution of a linear system first
and then project the solution on set $\mathcal{D}$.
Though this strategy seems to be efficient in practice \cite{bao2016dictionary}, it lacks theoretical guarantee.
Firstly, $\D^{t+1}$ calculated by INV is not an exact solution of (\ref{Exp:inexact_D}).
Secondly, it is computed without measuring the inexactness.
So applying INV sometimes creates oscillations during iterations (Figure \ref{Fig:converge_synthetic} (f)) and the performances of INV are unstable especially in real-world applications (see the experimental results in Section \ref{Sec:Exp:Real}).
Thus we do not recommend using it in practice.

%In the end, we compare the computational cost of PALM, IPAD-ADMM and INV at every iteration.
%Since these three algorithms share the same updates of $\W^{t+1}$, the only difference of the computational cost lies in the update of $\D^{t+1}$.
%We list the computational cost of updating $\D^{t+1}$ for only once.
%The number of the dominant operations of PALM is $\mathcal{O}(m^3+mnp+mn)$.
%Calculating $\D^{t+1}$ by IPAD-ADMM will cost $\mathcal{O}(m^3n+m^2p+mnp+mn)$.
%The last algorithm, INV has $\mathcal{O}(m^3n+m^2p+mnp+mn)$ calculations.
%
%Though INV seems to take more operations than PALM, the experimental results in Table \ref{Tab:Synthetic} do not match this expectation: INV converges with more steps but uses less time.
%We should mention that the calculations $\mathcal{O}(m^3n)$ comes from using Gauss-Elimination to obtain an inverse of matrix.
%However, the inversion process in Matlab employ more efficient methods.
%This is the reason for causing the contradiction between theory and practice.

\begin{figure}[!t]\label{FIG:con1}
\hspace{0.0cm}
\begin{minipage}[b]{0.49\linewidth}%4
  \centering
  \centerline{\includegraphics[width=1.1\textwidth]{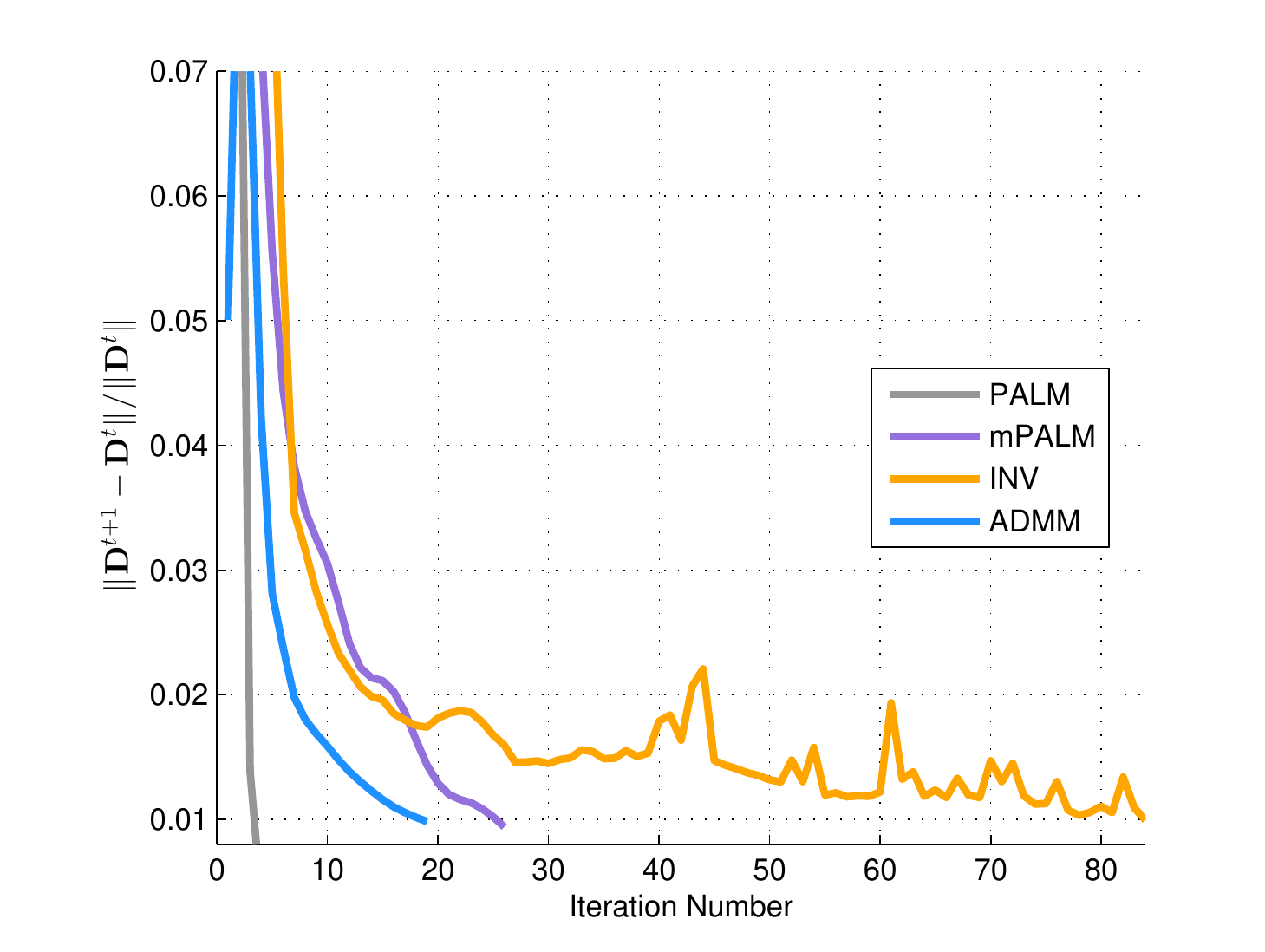}}
 %\vspace{1.5cm}
 %\vspace{-0.1cm}
  \centerline{(a)  ``Hill512'', $\sigma=15$}\medskip
\end{minipage}
\begin{minipage}[b]{0.49\linewidth}%4
  \centering
  \centerline{\includegraphics[width=1.1\textwidth]{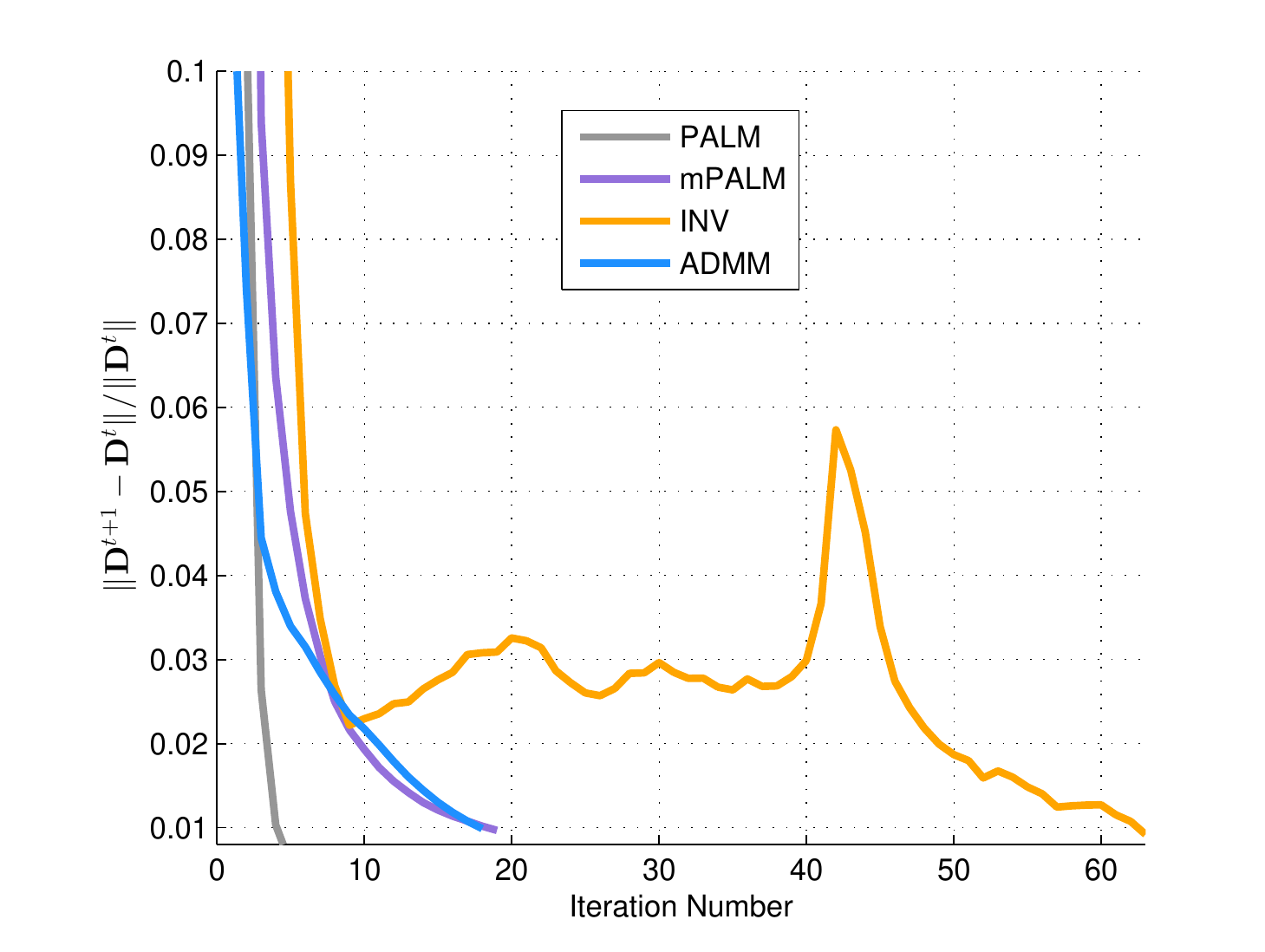}}
 %\vspace{1.5cm}
 %\vspace{-0.1cm}
  \centerline{(b)  ``Child512'', $\sigma=30$}\medskip
\end{minipage}
\caption{Comparing the convergence performances with various algorithms on two images.}
\end{figure}

\begin{figure}[!t]\label{FIG:con2}
\hspace{0.0cm}
\begin{minipage}[b]{0.49\linewidth}%4
  \centering
  \centerline{\includegraphics[width=1.1\textwidth]{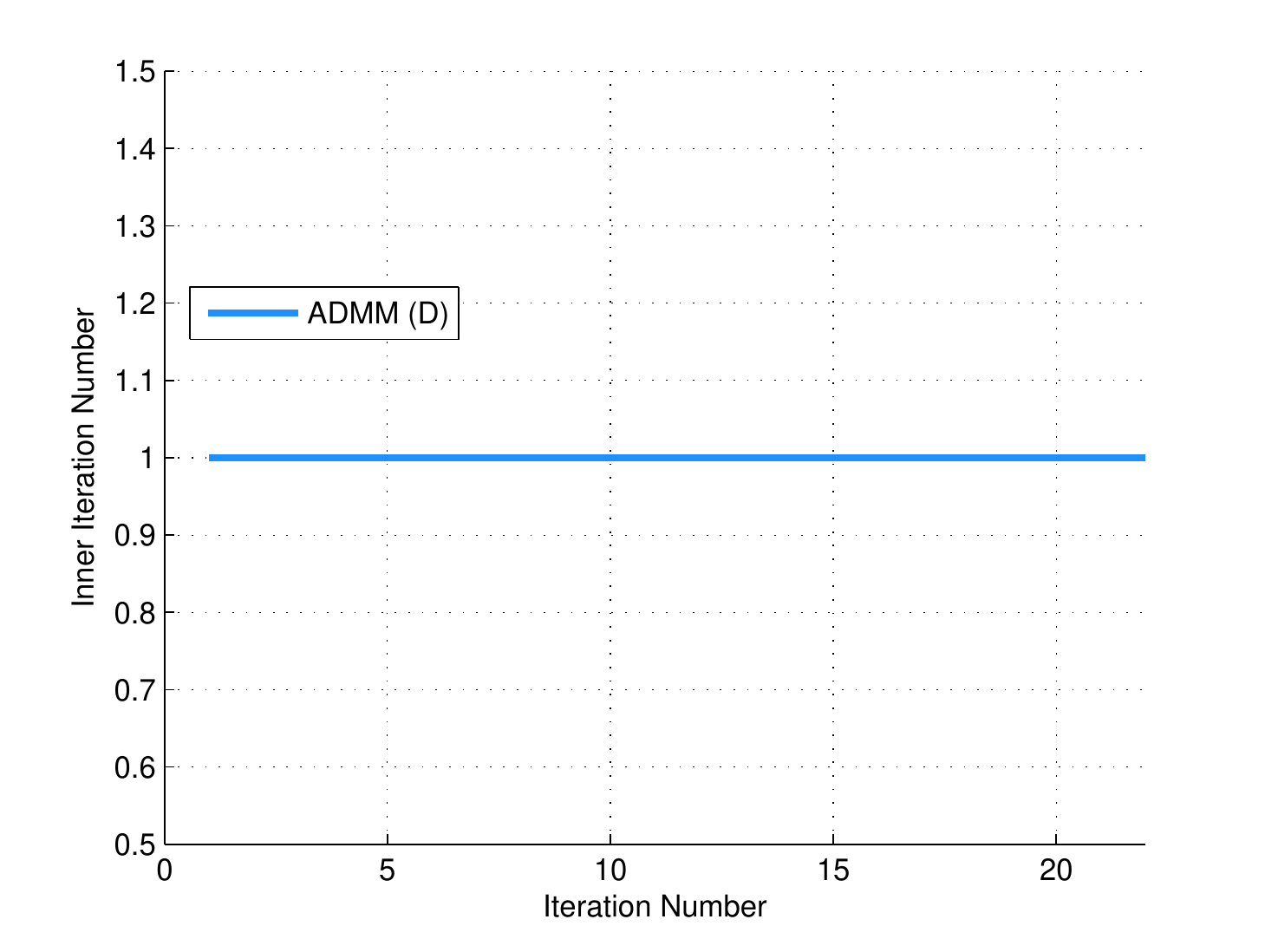}}
 %\vspace{1.5cm}
 %\vspace{-0.1cm}
  \centerline{(a)  ``Couple512'', $\sigma=20$}\medskip
\end{minipage}
\begin{minipage}[b]{0.49\linewidth}%4
  \centering
  \centerline{\includegraphics[width=1.1\textwidth]{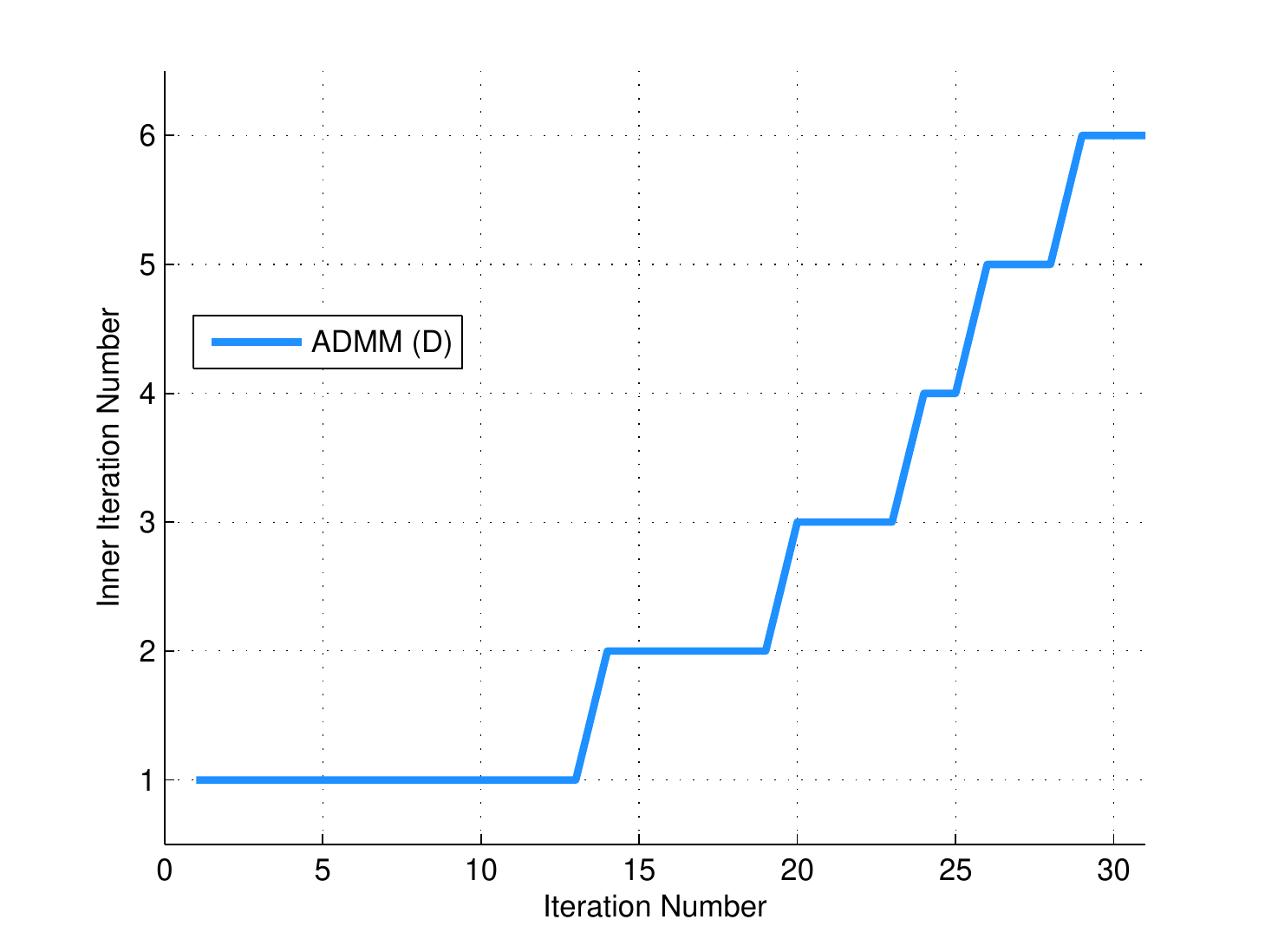}}
 %\vspace{1.5cm}
 %\vspace{-0.6cm}
  \centerline{(b)  ``Lena512'', $\sigma = 25$}\medskip
\end{minipage}
\caption{Inner iterations of IPAD-ADMM on two examples. }
\end{figure}

\subsection{Real-world Data}\label{Sec:Exp:Real}

We apply IPAD to real-world data on image denoising problem \cite{elad2006image,Chen2013Group} and compare PALM \cite{bolte2014proximal}, mPALM \cite{bao2014l0}, INV \cite{bao2016dictionary} with our IPAD-ADMM for solving this real-world application.
All the algorithms for comparison terminate when reaching $\|\D^{t+1}-\D^{t}\|/\|\D^t\|<1e^{-2}$ and are demonstrated on 7 widely-used images.
The patches in each image, of size $8\times 8$, are regularly sampled in an overlapping manner.
The noisy images are obtained by adding Gaussian randomly noises with level $\sigma = 15, 20, 25, 30$.

As shown in Table \ref{Tab:Real-world}, PALM seems to converge quickly but get bad recovered results.
However, the truth is: the large upper bound $\|(\D^{t+1})^{\top}\D^{t+1}\|_2$ of the Lipschitz constant emphasizes the function of the proximal term.
So it causes tiny differences between $\D^{t+1}$ and $\D^{t}$. Thus, PALM does not converge when reaching the stopping criterion; on the contrary, it converges quite slow.
For the failure of using PALM, we adopt the strategy used in \cite{bao2014l0}, which regards the problem (\ref{Exp:DL_model}) as a multi-block problem: solving $\{\bd_i\}_{i=1}^m$ separately by PALM.
We name their algorithm mPALM and show the results in Table \ref{Tab:Real-world}.
This time, mPALM converges to an acceptable result when reaching the stopping criterion.

We in Fig. \textcolor{red}{2} present comparison results on convergence performances: the vibration of INV \cite{bao2016dictionary} causes more iterations than IPAD-ADMM.
We also select two examples in Fig. \textcolor{red}{3} to present the inner iteration numbers of our inexact algorithm.
It can be seen from the figure that the inner iteration behaviors may be in totally different style; it depends on the input data and algorithm parameters.
In addition, we also some visual comparisons in Fig. \textcolor{red}{4} and \textcolor{red}{4} to show that our IPAD-ADMM performs more stable and efficient for the real-world applications.

Finally, we give comparisons between the state-of-the-art KSVD technique \cite{elad2006image} and our proposed IPAD-ADMM in Table \ref{Tab:KSVD}.
First we want to point it out that KSVD is designed for solving a quasi-$\ell_0$ (not the exact $\ell_0$ penalty) problem with the usage of OMP \cite{tropp2007signal}.
We can see that though our PSNR values are slightly less than KSVD, we use less than a tenth of the time of KSVD.
Thus, our IPAD is a fast algorithm with flexible inner iterations for solving the problem SDL with $\ell_0$ penalty in practice.

\begin{figure}[!t]
\begin{minipage}[b]{0.32\linewidth}%4
  \centering
  \centerline{\includegraphics[width=1\textwidth]{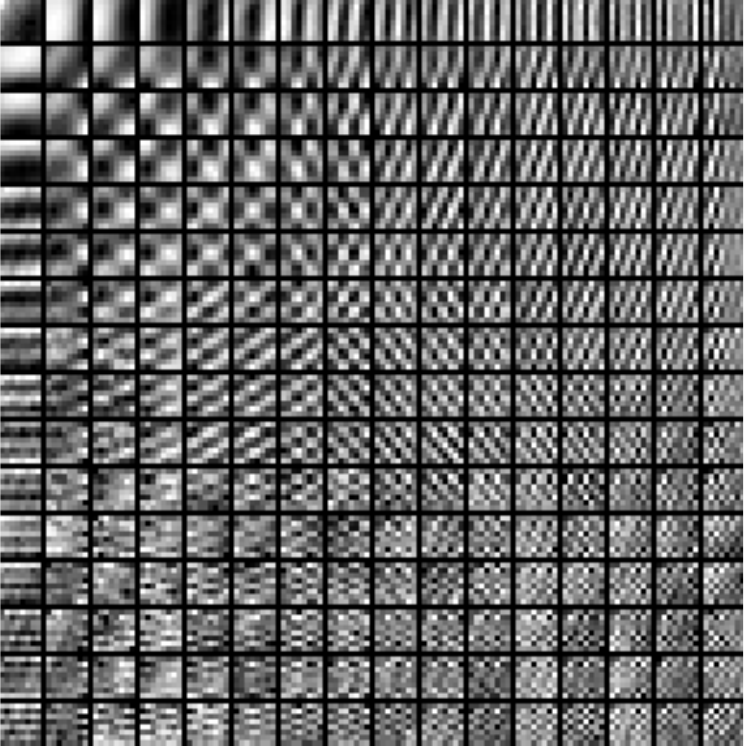}}
 %\vspace{1.5cm}
 %\vspace{-0.1cm}
  \centerline{(a)  mPALM \cite{bao2014l0}}\medskip
\end{minipage}
%\hspace{0.1cm}
\begin{minipage}[b]{0.32\linewidth}%4
  \centering
  \centerline{\includegraphics[width=1\textwidth]{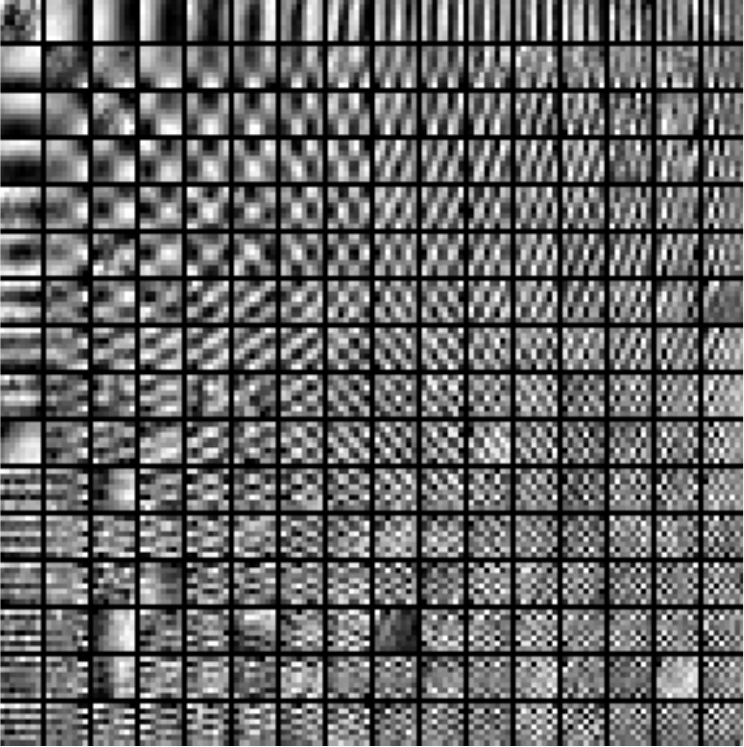}}
 %\vspace{1.5cm}
 %\vspace{-0.6cm}
  \centerline{(b) INV \cite{bao2016dictionary}}\medskip
\end{minipage}
%\hspace{0.1cm}
\begin{minipage}[b]{0.32\linewidth}%4
  \centering
  \centerline{\includegraphics[width=1\textwidth]{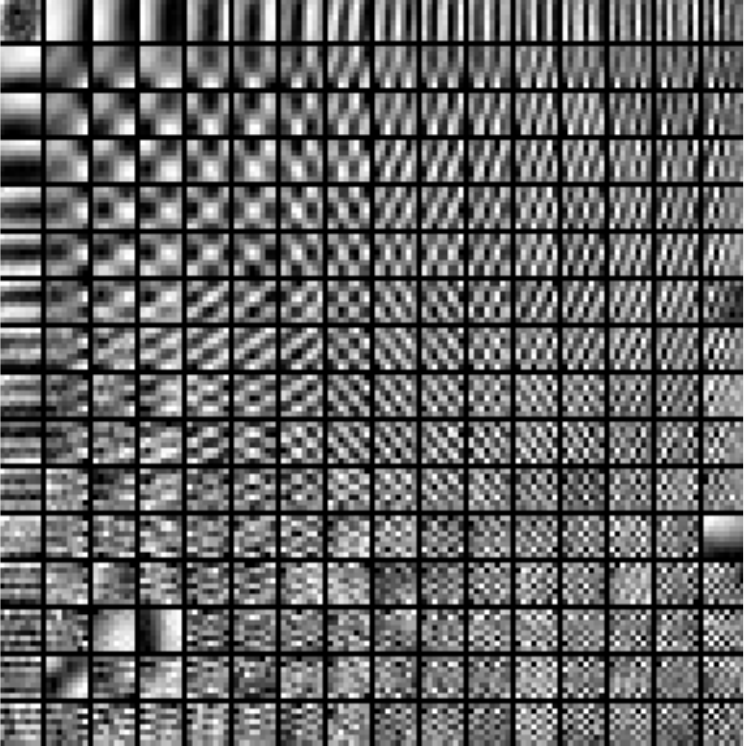}}
 %\vspace{1.5cm}
 %\vspace{-0.6cm}
  \centerline{(c) IPAD-ADMM}\medskip
\end{minipage}
\vspace{-0.2cm}
\label{FIG:dic}
\caption{The dictionaries learned by mPALM \cite{bao2014l0}, INV \cite{bao2016dictionary} and our IPAD-ADMM on ``Barbara512'' with $\sigma = 20$.}
\end{figure}

\begin{figure}[!t]
	\hspace{0.2cm}
	\begin{minipage}[b]{0.44\linewidth}%4
		\centering
		\centerline{\includegraphics[width=1\textwidth]{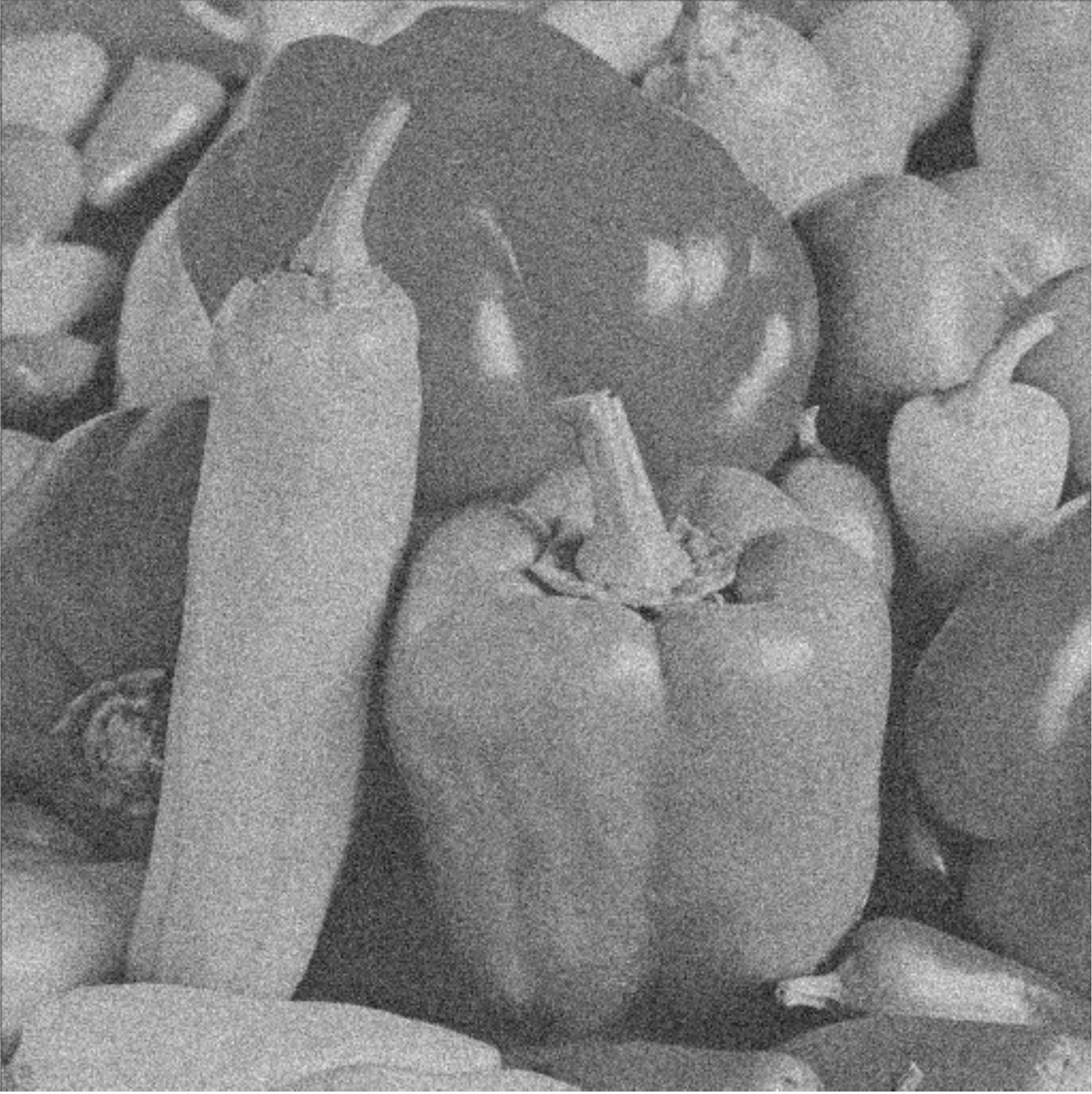}}
		%\vspace{1.5cm}
		%\vspace{-0.1cm}
		\centerline{(a)  Noisy image}\medskip
	\end{minipage}
	\hspace{0.1cm}
	\begin{minipage}[b]{0.44\linewidth}%4
		\centering
		\centerline{\includegraphics[width=1\textwidth]{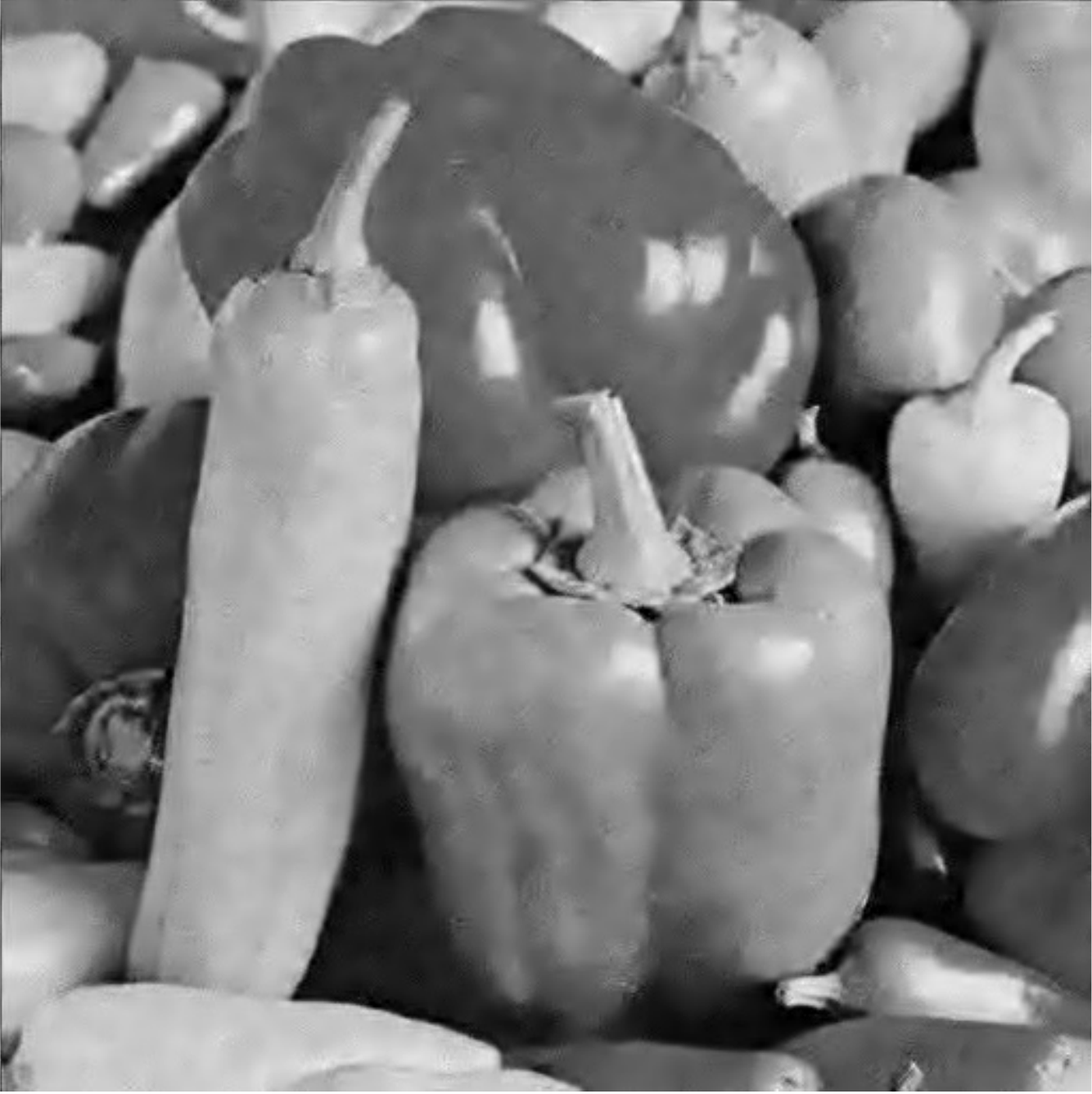}}
		%\vspace{1.5cm}
		%\vspace{-0.6cm}
		\centerline{(b) Recovered image}\medskip
	\end{minipage}
	\vspace{-0.1cm}
	\caption{Illustrating image denoising performance on an example image. (a) noisy image with $\sigma=30$, (b)  image recovered by IPAD-ADMM.}
\end{figure}\label{FIG:image4}

\section{Conclusions}

This paper provided a fast optimization framework to solve the challenging nonconvex and nonsmooth
programmings (NNPs) in vision and learning societies. Different from most existing solvers, which always fix their
updating schemes during iterations, we showed that under some mild conditions, any numerical algorithms can be incorporated
into our general algorithmic framework and the convergence of the hybrid iterations can always be guaranteed.
We conjectured that our theoretical results are the best we can ask for unless further assumptions are made on the general NNPs in \eqref{eq:primal}. Numerical evaluations on both synthetic data and real images demonstrated promising
experimental results of the proposed algorithms.

%In this paper, we propose a general and flexible algorithm framework named IPAD for solving NNMF problems, which achieves the best convergence result for non-convex and non-smooth problems.
%Different from previously used inexact strategies, IPAD method gives rigid implementations of the stopping criteria, which is more robust than the previously used ones in practice.
%Moreover, we verify from the experimental results that IPAD is efficient and effective for solving a SDL problem with $\ell_0$ penalty.
%
%Since our IPAD method can be straightforwardly extended to a general multi-block non-convex and non-smooth problem, it can be applied to a wider range of applications beyond NNMF, like discriminative learning problems, optimizations on manifolds and tensor decompositions.
%Moreover, the iterative process of IPAD can be more improved: a random version of IPAD is welcomed to avoid local optimal; a hybrid form of IPAD is more flexible in practice just as claimed in the last section.
%At the moment, considering the generality and flexibility of IPAD method, we are focusing on discovering its connections with some frameworks in neural network; and then try to optimize those network structures to obtain more robust results.

\section{Appendix A: proof of Proposition \ref{prop}}
\begin{proof}
From the Alg. \ref{Alg:iPAM2}, at the $t$-th iteration, suppose that an inner iteration scheme is conducted by $K$ time, and denote its current iterative solution as $\x^{t,K}$.
Then from the calculations in Eq. (\ref{Alg_Imp:extra_variables}), we can deduce the following equalities.
\begin{equation}\label{eq:xxx}\small
\begin{aligned}
\widetilde{\x}^{t,K}=&\mathrm{prox}^{1}_{f}(\x^{t,K} - \nabla_{\x}H(\x^{t,K}, \y^{t-1}) - \eta_1^{t-1}(\x^{t,K}-\x^{t-1}))\\
=&\mathrm{prox}^{1}_{f}(\widetilde{\x}^{t,K} - \nabla_{\x}H(\widetilde{\x}^{t,K}, \y^{t-1}) - \eta_1^{t-1}(\widetilde{\x}^{t,K}-\x^{t-1})\\
+&(1-\eta_1^{t-1})(\widetilde{\x}^{t,K}-\x^{t,K})\\
+&\nabla_{\x}H(\x^{t,K}, \y^{t-1})-\nabla_{\x}H(\widetilde{\x}^{t,K}, \y^{t-1})).
\end{aligned}
\end{equation}
Once the error $\e_x^{t,K}$ satisfies the inexact condition Crit \ref{crit}, then $\widetilde{\x}^{t,K}$ will be regarded as the solution of the $t$-th iteration.
That is, by substituting the notation $\widetilde{\x}^{t,K}$ with $\x^t$ in Eq. (\ref{eq:xxx}), thus we get
$$
\x^{t} =\mathrm{prox}^{1}_{f}(\x^{t} - \nabla_{\x}H(\x^{t}, \y^{t-1}) - \eta_1^{t-1}(\x^{t}-\x^{t-1})+\e_x^{t}).
$$
The above deductions can be similarly extended to the case of $\y^t$.
Together with Eq. (\ref{eq:xx}), we have
\begin{equation}\label{eq:xxxx}
\x^{t} = \mathrm{prox}^{1}_{f}(\sv_{x}^{t}+\e_x^{t}), \ \  \y^{t} = \mathrm{prox}^{1}_{g}(\sv_{y}^{t}+\e_y^{t}).
\end{equation}
From the definition of $\mathrm{prox}$ mapping, Eq. (\ref{eq:xxxx}) is equal to
\begin{equation}
\begin{aligned}
\e_x^{t} &= \mathbf{g}_x^{t} + \nabla_{\x} H(\x^{t}, \y^{t-1}) + \eta_1^{t-1}(\x^{t}-\x^{t-1}),\\
\e_y^{t} &= \mathbf{g}_y^{t} + \nabla_{\y} H(\x^{t}, \y^{t}) + \eta_2^{t-1}(\y^{t}-\y^{t-1}).
\end{aligned}
\end{equation}
where $\mathbf{g}_x^{t+1} \in \partial f(\x^{t+1})$ and $\mathbf{g}_y^{t+1} \in \partial g(\y^{t+1})$.
The above equalities are exactly the first-order optimality conditions of (\ref{iPAM:iex_sequence}) by regarding $\e_x^t$ and $\e_y^t$ as the inexactness.
$\blacksquare$
\end{proof}

%{\small
%\bibliographystyle{ieee}
%\bibliography{egbib}
%}

\end{document}